\pdfoutput=1

\documentclass[11pt]{article}

\usepackage[preprint]{acl}

\usepackage{times}
\usepackage{latexsym}

\usepackage[T1]{fontenc}

\usepackage[utf8]{inputenc}

\usepackage{microtype}

\usepackage{inconsolata}

\usepackage{graphicx}
\usepackage{soul,xcolor}
\usepackage{todonotes}
\usepackage{paralist}
\usepackage{lipsum}

\usepackage{xspace}
\renewcommand{\arraystretch}{1.2}
\usepackage{multirow}

\newcommand{\dataset}{\textsc{LoFTI}\xspace}

\title{\dataset: Localization and Factuality Transfer to Indian Locales}

\author{
 \textbf{Sona Elza Simon\textsuperscript{$\ddagger$}},
 \textbf{Soumen Kumar Mondal,\textsuperscript{$\ddagger$}},
 \textbf{Abhishek Singhania\textsuperscript{$\S$}},
 \\
 \textbf{Sayambhu Sen\textsuperscript{$\S$}},
 \textbf{Preethi Jyothi\textsuperscript{$\ddagger$}}
\\
\\
 \textsuperscript{$\ddagger$}
Indian Institute of Technology Bombay, Mumbai, India,\\
 \textsuperscript{$\S$}Amazon Alexa 
 \\
\texttt{\normalsize{\{sona.simon,23m2157,pjyothi\}@iitb.ac.in, \{mrabhsin,sensayam\}@amazon.com}
 }
}

\usepackage{multirow}
\usepackage{tabularx}
\usepackage{tablefootnote}
\usepackage[most]{tcolorbox}

\begin{document}

\maketitle
\begin{abstract}
Large language models (LLMs) encode vast amounts of world knowledge acquired via training on large web-scale datasets crawled from the internet. However, these datasets typically exhibit a geographical bias towards English-speaking Western countries. This results in LLMs producing biased or hallucinated responses to queries that require answers localized to other geographical regions. In this work, we introduce a new benchmark named \dataset (Localization and Factuality Transfer to Indian Locales) that can be used to evaluate an LLM's localization and factual text transfer capabilities. \dataset consists of factual statements about entities in source and target locations; the source locations are spread across the globe and the target locations are all within India with varying degrees of hyperlocality (country, states, cities). The entities span a wide variety of categories. We use \dataset to evaluate Mixtral, GPT-4 and two other Mixtral-based approaches well-suited to the task of localized factual transfer. We demonstrate that \dataset is a high-quality evaluation benchmark and all the models, including GPT-4, produce skewed results across varying levels of hyperlocality.
\end{abstract}

\section{Introduction}

\begin{figure}[t]
    \centering
    \includegraphics[width=\linewidth]{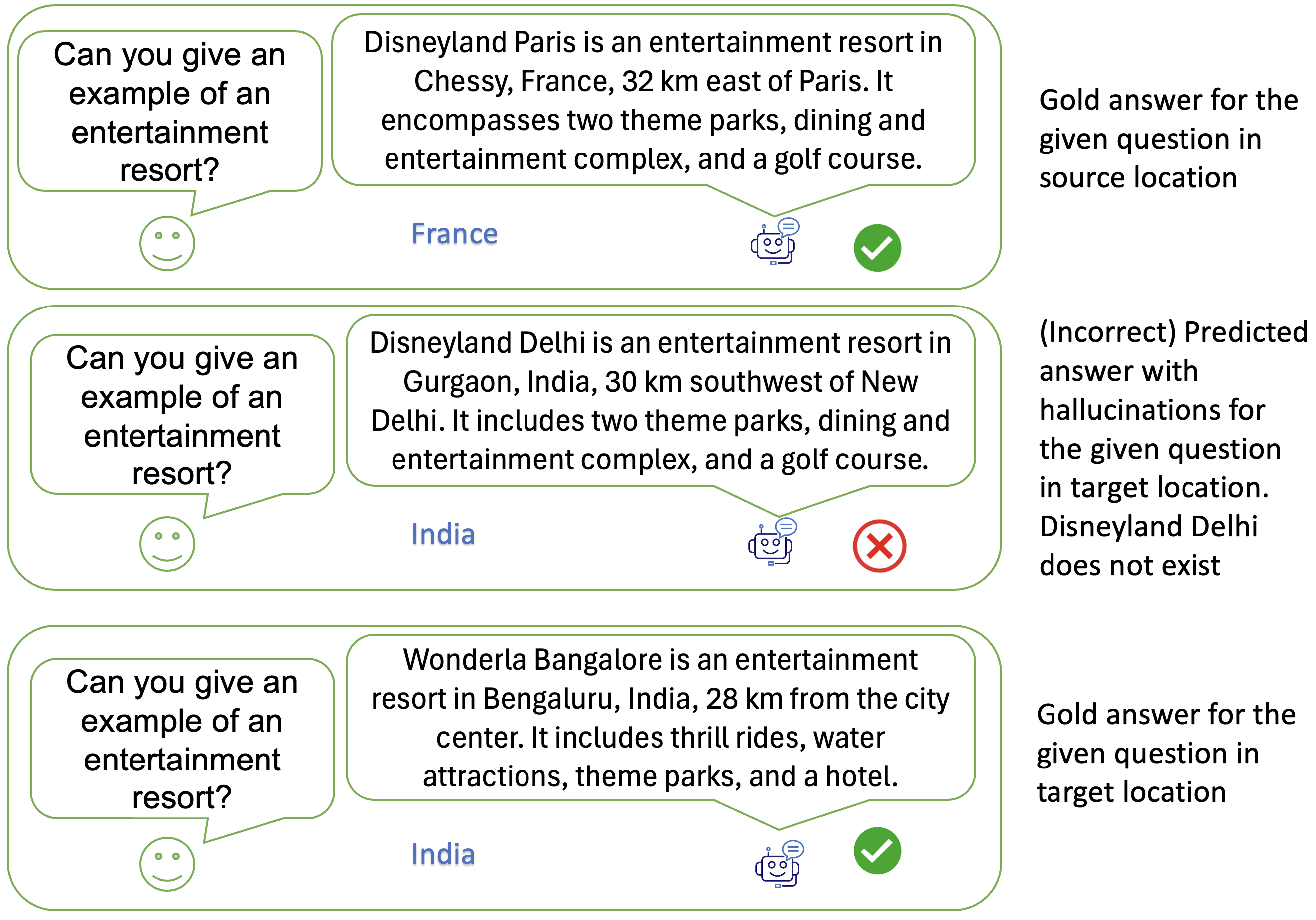}
    \caption{Illustration of LLM’s localized factual text transfer capabilities.}
    \label{fig:image}
\end{figure}

Large language models (LLMs) are proficient in text generation and are also extensive repositories of world knowledge, owing to their pretraining and fine-tuning on vast and diverse internet data. This suggests that LLMs might be effective at transferring factual knowledge across geographical locations. They can generate localized text in a given target location by transferring from a reference text in a source location. However, there is no existing benchmark that helps assess this specific form of localization and fact-driven transfer. Benchmarks that measure LLMs' ability to understand cultural concepts and their transference across geographical regions are steadily emerging in recent work~\citep{li2024culturellm, li2024culturegen, li2024culturepark,rao2024normad}. We argue that it is also important to evaluate the ability of models to transfer factual knowledge across geographical regions. Figure~\ref{fig:image} illustrates this point by showing two use-cases: 1) Generating a localized response given a common question that can be asked across locations and, 2) accurate factuality transfer from one locale to another. 

In this work, we introduce a new evaluation benchmark called \dataset (Localization and Factuality Transfer for Indian Locales). Notable features of \dataset are:
\begin{asparaitem}
\item It contains factual statements in source and target locations involving source and target entities.
\item The statements are grounded in various source locations across the globe, while all the target locations are in India.
\item The target locations are at different levels of hyperlocality namely specific to India as a whole, or specific to states and cities within India.
\item The entities in the statements span a diverse set of categories including food, sports, nature, etc. 
\item Each parallel set of statements is accompanied by (one or more) common questions that can be answered at any location.
\end{asparaitem}

The motivation behind creating \dataset stems from the lack of comprehensive multi-locale (and multilingual data) on the internet, which is essential for both training and evaluating LLMs. Simple translations of English datasets are inadequate because they predominantly feature Western entities and facts, introducing biases that are irrelevant or inaccurate for non-Western locales. \dataset can be used as a benchmark to help improve LLMs on factuality transfer in English from reference to target locations. Once we have high-performing LLMs on this task in English, we could potentially create multilingual factual data using direct translations of the target text into languages specific to the target locations. \dataset can also be used to benchmark multilingual/multi-locale LLMs by evaluating their performance on localized question answering with different context locations. 


In this work, we define three different metrics to evaluate the quality of both localization and factuality transfer on \dataset. We evaluate the performance of powerful open-source (Mixtral) and closed-source models (GPT-4) on \dataset. We also develop two variants of Mixtral that leverage external sources of evidence to significantly improve performance on all three metrics. While GPT-4 is expectedly superior in performance compared to all Mixtral variants, it shows degradation in performance across target locations of varying hyperlocality, thus revealing clear gaps in coverage across geographical regions. 
We publicly release \dataset under the Apache 2.0 license \footnotemark{}.
\footnotetext{The \dataset dataset and codebase are available at: \url{https://huggingface.co/datasets/sonasimon/LoFTI} 
\url{https://github.com/csalt-research/LoFTI}}



\section{Methodology for Dataset Creation}
\begin{figure}[h!]
    \centering
    \includegraphics[width=0.9\linewidth]{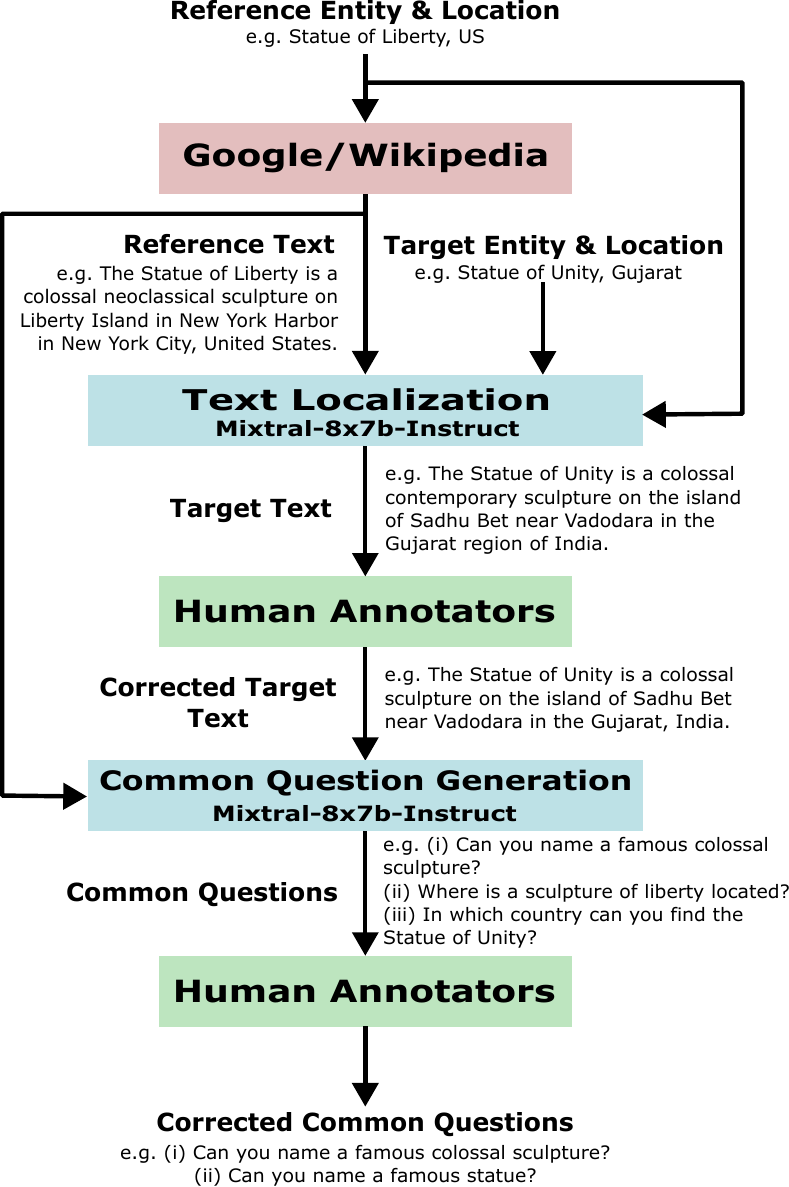}
    \caption{Illustration of the dataset creation pipeline with an example.}
    \label{fig:pipeline}
\end{figure}

Figure \ref{fig:pipeline} describes the overall dataset creation pipeline with the help of an example. Next, we outline the details of each step in the dataset creation process.

\subsection{Generation of Entity-pairs}
For dataset creation, we compile pairs of entities $(e_{\text{ref}}, e_{\text{tar}})$, where $e_{\text{ref}}$ is an entity from a reference location outside India and $e_{\text{tar}}$ is an entity from India that serves as a suitable substitute for $e_{\text{ref}}$. These pairs are curated by human annotators and cover diverse categories and hyperlocal regions.
\looseness=-1

\subsection{Reference Text Generation}
Given the reference entity $e_{\text{ref}}$, a fact-based reference text $T_{\text{ref}}$ is obtained from the entity's description on the internet. We use the Google API Client or Wikipedia for this purpose. If no entity description is found from these sources, human annotators are tasked with providing the reference text.

\subsection{Text Localization} 
Given a reference text $T_{\text{ref}}$ and a target entity $e_{\text{tar}}$ (paired with $e_{\text{ref}}$) from a target location $L_{\text{tar}}$, text localization aims to generate a  target text $T_{\text{tar}}$ localized to $L_{\text{tar}}$ that retains the stylistic and semantic features of $T_{\text{ref}}$. This process involves localizing the entities and facts present in $T_{\text{ref}}$ while ensuring factual correctness. For text localization, we employ the \emph{Mixtral-8x7b-instruct-v0.1.Q4\_K\_M} model. Given the target location $L_{\text{tar}}$, target entity $e_{\text{tar}}$, and the reference text $T_{\text{ref}}$, we prompt the Mixtral model to generate the localized target text $T_{\text{tar}}$. The prompt used for text localization is given in Figure \ref{fig:prompt_loc}.
%
%
    \begin{table*}[t!]
    \centering
    \scriptsize
    \begin{tabularx}{\textwidth}{lclclc}
        \hline
        \multicolumn{2}{l}{\textbf{\dataset Dataset Details}} & \multicolumn{2}{c}{} & \multicolumn{2}{c}{} \\ \hline
        No. of entity pairs & 1100 & No. of entities from US/Europe & 651 & No. entities with high cardinality & 835\\ 
        No. of categories & 99 & No. of entity pairs from other places & 449 & No. entities with low cardinality & 265\\ 
        No. of entities with hyperlocal score = 1 & 369 & No. of entities with hyperlocal score = 2 & 391 & No. of entities with hyperlocal score = 3 & 34 \\ \hline
    \end{tabularx}
    \setlength{\tabcolsep}{3.1pt} 
    \begin{tabularx}{\textwidth}{p{0.8cm}p{1cm}p{1cm}p{1cm}p{2.25cm}p{1cm}p{0.75cm}p{0.75cm}p{1cm}p
        {2cm}p{2cm}}
        \textbf{Example: } &&&&&&&&& \\ \hline
        \textbf{Region} & \textbf{Category} & \centering \textbf{Reference Location} & \centering \textbf{Reference Entity} & \centering \textbf{Reference Text} & \centering \textbf{Target Location} & \centering \textbf{High Cardinality} & \textbf{Hyperlocal Score} & \centering \textbf{Target Entity} & \centering \textbf{Target Text} & \multirow{2}{*}{\textbf{Common Questions}} \\ \hline
        \centering US/ Europe & Monument & \centering US & \centering Statue of Liberty & The Statue of Liberty is a colossal neoclassical sculpture on Liberty Island in New York Harbor in New York City, United States. & \centering Gujarat & \centering Yes & \centering 2 & \centering Statue of Unity & The Statue of Unity is a colossal sculpture on the island of Sadhu Bet near Vadodara in the Gujarat, India. & (i) Can you name a famous colossal sculpture?\newline(ii) Can you name a famous statue? \\ \hline
    \end{tabularx}
    \caption{The statistics of \dataset dataset and an example with all its metadata.}
    \label{table:dataset_example}
\end{table*}
\begin{figure*}[t!]
    \centering
    \includegraphics[width=\textwidth]{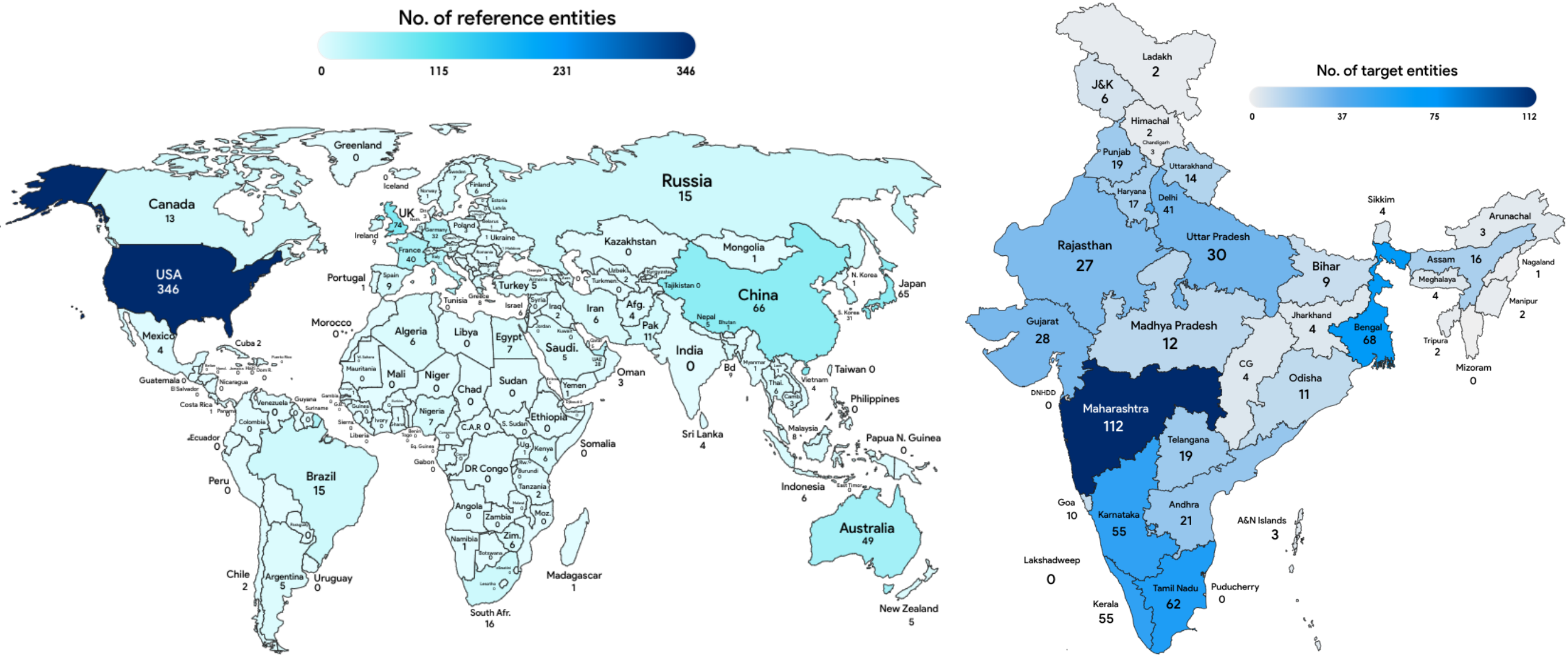}
    \caption{\centering Illustrates the global distribution of the reference entities and the spread of target entities in India.}
    \label{fig:map}
\end{figure*}
\subsection{Common Question Generation}
In addition to the reference and target text pairs, \dataset  also contains questions that capture common aspects shared by $T_{\text{ref}}$ and  $T_{\text{tar}}$. Given a pair of text $(T_{\text{ref}}, T_{\text{tar}})$, we generate these questions by identifying shared properties or descriptions of the entities mentioned in the text pairs. We use few-shot prompting on \emph{Mixtral-8x7b-instruct-v0.1.Q4\_K\_M} model for common question generation and the prompt used is given in Figure \ref{fig:prompt_cq}. 

\subsection{Human Annotators}
To ensure the correctness of the \dataset dataset, all the generations were carefully checked by human annotators at each stage. These annotators represent diverse demographics and have knowledge about samples from different geographic and hyperlocal regions. Each sample undergoes verification by three annotators. Guidelines used by the human annotators at each stage are detailed in Appendix \ref{sec:annotation}.

\section{Properties of \dataset Dataset}

    \dataset consists of factual texts that are localized from a non-Indian reference location to a location in India. The reference locations are spread across the globe, mainly in USA/Europe. The target locations are spread across India covering different regions. Figure \ref{fig:map} shows the distribution of entities across reference and target locations. 
    
    Table~\ref{table:dataset_example} presents salient statistics of \dataset and an example with all its metadata detailed below.
    \begin{asparaitem}
        \item \textbf{Region}: The region of the reference location. 
        \item \textbf{Category}: The category of the entity in the factual text.
        \item \textbf{Reference Location}: A non-Indian location.
        \item \textbf{Reference Entity}: An entity from the reference location.
        \item \textbf{Reference Text}: Factual text about the reference entity.
        \item \textbf{Target Location}: A location in India.
        \item \textbf{True Target Entity}: An example of a correct localization of the reference entity in the target location.
        \item \textbf{True Target Text}: A localized factual text of the true target entity.
        \item \textbf{Hyperlocal Score}: The degree of hyperlocality within the Indian context. The dataset includes three hyperlocality scores: 1, 2, and 3. These scores correspond to the target locations `India,' `any state in India,' and `any city in India,' respectively.
        \item \textbf{High Cardinality}: Cardinality denotes the potential count of replaceable entities for the reference entity within the target location. A high cardinality suggests there are many such replaceable entities. This feature is denoted by 'yes' or 'no' values.
        \item \textbf{Common Questions}: Questions extracted from the reference and the target texts.
    \end{asparaitem}
      
    \paragraph{Category Distribution.} The dataset consists of 99 unique categories which can be grouped into 10 domains namely Entertainment, Buildings/Monuments/Companies, Food \& Lifestyle, Professions, Nature, Finance \& Economy, Sports, Incidents, Places \& Landmarks, and Others. The category clusters and the category distribution are shown in Table \ref{table:categories} and Figure \ref{fig:count_category},  respectively.   

\section{Evaluation Metrics}

    \subsection{Entity Correctness}
    To evaluate entity correctness of a generated target text, the human annotator checks if the entity present in the target text is correctly localized to a target location given the reference entity in the reference text. Note that there can be multiple correct localized entities for a given target location. If the entity localization is correct, then a score of 1 is assigned, else it is 0. Thus, for each generated target text, $T^i$, a binary score $E^i$ is assigned. Across $N$ generated target text sequences, the entity correctness metric is computed as $\textbf{EC} = \frac{1}{N} \sum_{i=1}^{N} E^i$.

    \subsection{Common Question Correctness}
    For each target text with $\text{EC} = 1$, the common questions present in \dataset are further used to evaluate the localization capability of the model. Human evaluators check if the target text correctly answers the common questions given the target location. Each question is evaluated separately and they assign a binary score of 1 if it is answered correctly, else it is 0.
    	
    For a generated target text $T^i$, let the number of predefined common questions be $m_i$ and the binary scores for these questions be $\{C_j^i\}_{j=1}^{m_i}$. Then, the common question correctness metric across $N$ texts is calculated as $\textbf{CQ} = \frac{1}{\sum_{i=1}^N m_i} \sum_{i=1}^{N} \sum_{j=1}^{m_i} C_j^i$. This metric aggregates the scores across all questions for all target texts, providing an overall measure of the model’s effectiveness in generating contextually accurate and relevant responses.

    \subsection{Factual Correctness}
     For each target text $T^i$ with $\text{EC} = 1$,  the human annotator checks if every detail in the text is factually correct and provides a binary score $F^i=1$ if every fact is correct, else $F^i=0$. The factual correctness metric across $N$ texts is calculated as $\textbf{FC} = \frac{1}{N} \sum_{i=1}^{N} F^i$.

\section{Models and Approaches}

\subsection{Models}
We evaluate the performance of two state-of-the-art LLMs on \dataset: Mixtral~\citep{jiang2024mixtral} and GPT-4 \citep{openai2023gpt4}. The Mixtral-8x7B LLM \citep{jiang2024mixtral} is a pre-trained generative sparse mixture-of-experts model. It has a decoder-only architecture with its feedforward block selecting from a set of 8 distinct groups of parameters. 
For our analysis, we utilize the quantized Mixtral model \emph{Mixtral-8x7b-instruct-v0.1.Q4\_K\_M\footnotemark{}} with zero-shot prompting. Interestingly, we observed that few-shot prompting did not improve performance compared to the zero-shot setting and adding more localization examples appeared to confuse the model.
\footnotetext{The quantized Mixtral-8x7b models Q6 and Q8 gave similar performance to Q4.} We also evaluate the performance of the state-of-the-art GPT-4 model on \dataset. We use the same prompt for both Mixtral and GPT-4 (detailed in Appendix~\ref{sec: zero_shot_prompt}).

\subsection{Approaches}

\paragraph{Mixtral + RARR.} \label{sec:rarr} LLM generations, while fluent, are known to be prone to hallucinations and factual inaccuracies. To address this, \citet{gao2022rarr} proposed RARR (Retrofit Attribution using Research and
Revision), an attribution mechanism that leverages external evidence from the web to validate and edit LLM-generated text while aiming to maintain the original style of the output. We utilize RARR to factually correct the generations produced by Mixtral.

RARR consists of three modules: (i) Question Generation Module, (ii) Evidence Retrieval Module, and (iii) Editor Module. The Question Generation Module formulates questions from the text to be edited and the Evidence Retrieval Module queries these questions on the web for factual evidence. While querying, the target location of the text is appended to the start of each question to extract evidence relevant to that location. The retrieval module also checks if the text to be edited disagrees with the evidence. The Editor Module then utilizes all the disagreed evidence to make factual edits to the text. We employ the \emph{Mixtral-8x7b-instruct-v0.1.Q4\_K\_M} model in both the Question Generation and Editor Modules. As in the original RARR pipeline, we utilize Microsoft Bing for evidence retrieval. We adhere to the RARR pipeline, except for one detail. We aggregate all the evidence obtained for all the generated questions and make a single edit, whereas RARR makes edits for each question individually. We found that sequential editing increased the text context and disrupted the style. Making a single edit helped maintain the text length and style better.

\paragraph{Mixtral Revised.} To improve the factual accuracy of the Mixtral generations, we propose a revised version (henceforth referred to as Mixtral Revised). Motivated by RARR, we use the Question Generation and Evidence Retrieval Modules as discussed in Section \ref{sec:rarr}. However, we replace the Editor module with a Re-generation module which filters the evidence and re-generates the text using \emph{Mixtral-8x7b-instruct-v0.1.Q4\_K\_M} model. The evidences retrieved from the Evidence Retrieval Module are filtered to assess their relevance to the context and they are added to the localized text transfer prompt to obtain more factually correct re-generation. This approach focuses on improving the factual correctness of the entity generated by Mixtral while preserving the style.

All the prompts used in the two above-mentioned approaches are detailed in Appendix~\ref{sec:RARR} and~\ref{sec:mixtral_revised}.

\section{Experiments and Results}
\subsection{Comparison of Mixtral and GPT-4 Performance for Localized Text Transfer}
Table~\ref{table:mixtral_vs_gpt4} compares the performance of state-of-the-art models GPT-4 and Mixtral on \dataset using all three metrics that collectively help evaluate localization and factual transfer. We observe that GPT-4 significantly outperforms Mixtral in EC, FC, and CQ by 17\%, 14\%, and  27\%,  respectively. As hyperlocal scores increase, both models show a decline in accuracy, indicating the difficulty in recalling knowledge about these locales. Benefiting from training on diverse and larger datasets and tasks, GPT-4 surpasses Mixtral by achieving superior localization even at higher hyperlocal scores, resulting in more accurate outputs.
\begin{table}[t!]
    \centering
    \footnotesize
    \begin{tabular}[\textwidth]{ccccc}
    \hline
        \textbf{Model} & \textbf{\# Samples} & \textbf{EC} & \textbf{CQ} & \textbf{FC} \\ \hline
        Mixtral & 1100 & \textbf{0.63} & \textbf{0.50} & \textbf{0.35}	  \\ 
        Hyperlocal 1 & 369 & 0.72 & 0.58 & 0.41 \\ 
        Hyperlocal 2 & 391 & 0.63 & 0.49  & 0.38 \\ 
        Hyperlocal 3 & 340 & 0.54 & 0.43 & 0.25 \\ \hline
        GPT-4 & 1100 &  \textbf{0.80} & \textbf{0.64} &	\textbf{0.62}  \\ 
        Hyperlocal 1 & 369 & 0.85 & 0.71 & 0.67 \\ 
        Hyperlocal 2 & 391 & 0.80 & 0.63 & 0.63 \\ 
        Hyperlocal 3 & 340 & 0.75 & 0.59 & 0.56 \\ \hline
    \end{tabular}
    \caption{Comparison of Mixtral and GPT-4 Performance for Localized Text Transfer.}
    \label{table:mixtral_vs_gpt4}
    \vspace{-0.1cm}
\end{table}
\begin{table*}[h]
\centering
\scriptsize
\setlength{\tabcolsep}{3.75pt}
\renewcommand{\arraystretch}{1.4}
\begin{tabular}{p{5cm}p{1.2cm}p{4.4cm}p{3.8cm}}
\hline
\centering \textbf{Reference Sentence} & \centering \textbf{Target Location} & \centering \textbf{Mixtral} & \textbf{\hspace{3em}GPT-4} \\ \hline
The Reichenbach Falls are a waterfall cascade of seven steps on the stream called Rychenbach in the Bernese Oberland region of Switzerland. They drop over a total height of about 250 meters (820 ft). & \centering India & The Dudhsagar Falls is a multi-step waterfall cascade on the Mandovi river in the Indian state of Goa, dropping over a total height of about \textcolor{red}{310 meters (1017 ft)}. & The Jog Falls are a segmented waterfall on the Sharavathi River located in Karnataka, India. They drop from a height of about 253 meters (830 ft).\\
Masashi Kishimoto is a Japanese manga artist. His manga series, Naruto, which was in serialization from 1999 to 2014, has sold over 250 million copies worldwide in 46 countries as of May 2019. & \centering Maharashtra & \textcolor{red}{Padmashri Rahul Aware} is a renowned Marathi cartoonist. His comic series, 'Toofan Tai', which was in serialization from 2005 to 2016, has sold over 3 million copies within Maharashtra as of May 2021. & Anant Pai was an Indian comic book creator. His comic book series, Amar Chitra Katha, which was launched in 1967, has sold over 100 million copies worldwide in multiple languages.\\
Eric Otto Valdemar Lemming was a Swedish track and field athlete who competed at the 1900, 1906, 1908, and 1912 Olympics in a wide variety of events, which mostly involved throwing and jumping. & \centering Khandra & \textcolor{red}{Surendra Kumar Singh} was a renowned Indian athlete who competed in several Athletics competitions at the 1956, 1960, and 1964 Olympics, primarily excelling in events such as shot put and javelin throw. & \textcolor{red}{Milkha Singh} was an Indian track and field athlete from Khandra who competed in the 1956, 1960, and 1964 Olympics, specializing in the 400 meters and other sprint events.\\ \hline
\end{tabular}
\begin{flushleft}
    \scriptsize{\hspace{1em}\textcolor{red}{Red} represents the error in the text.}\\
\end{flushleft}
\caption{Examples from different hyperlocal levels to illustrate the limitations of Mixtral and GPT-4 for localized text transfer.}
\label{table:example_gpt_mixtral}
\end{table*}
Table \ref{table:example_gpt_mixtral} shows examples highlighting the limitations of both Mixtral and GPT-4 at different hyperlocality values. For the example from hyperlocal score=1 (India), we observe that both models localize well but Mixtral tends to make errors in the factual details (e.g., the height of the waterfall). For hyperlocal score=2 (Maharashtra), Mixtral tends to hallucinate and creates an imaginary entity (``Padmashri Rahul Aware") while GPT-4 localizes correctly. For hyperlocal score=3 (Khandra), both models fails to localize the reference entity ``Eric Otto Valdemar Lemming" correctly. Mixtral returns an entity from a different category and location (``Surendra Kumar Singh" is a politician from Madhya Pradesh), while GPT-4 returns an entity from the correct category but a different location (``Milkha Singh" is a track-and-field athlete from Chandigarh). 
\begin{table*}[t!]
\centering
\footnotesize
\setlength{\tabcolsep}{2.9pt} 
\begin{tabular}{c|cccc|cccc|cccc|cccc}
\hline
\multicolumn{1}{c|}{\textbf{Models}} & \multicolumn{4}{c|}{\textbf{Mixtral}} & \multicolumn{4}{c|}{\textbf{Mixtral + RARR}} & \multicolumn{4}{c|}{\textbf{Mixtral Revised}} & \multicolumn{4}{c}{\textbf{GPT-4}} \\ \hline
\textbf{Hyperlocal} & \textbf{Overall} &\textbf{1} & \textbf{2} & \textbf{3} & \textbf{Overall} & \textbf{1} & \textbf{2} & \textbf{3} & \textbf{Overall} & \textbf{1} & \textbf{2} & \textbf{3} & \textbf{Overall} & \textbf{1} & \textbf{2} & \textbf{3} \\ \cline{1-17} 

\multicolumn{17}{c}{\textbf{Human Evaluation}} \\ \hline
\textbf{EC} & 0.60 & 0.70 & 0.63 & 0.45 & 0.65 & 0.77 & 0.65 & 0.48 & 0.66 & 0.78 & 0.65 & 0.52 & 0.81 & 0.83 & 0.83 & 0.75 \\ 
\textbf{CQ} & 0.48 & 0.54 & 0.53 & 0.36 & 0.57 & 0.68 & 0.61 & 0.39 & 0.56 & 0.64 & 0.57 & 0.45 & 0.65 & 0.68 & 0.66 & 0.60 \\ 
\textbf{FC} & 0.35 & 0.38 & 0.44 & 0.20 & 0.48 & 0.60 & 0.49 & 0.31 & 0.51 & 0.61 & 0.53 & 0.34 & 0.63 & 0.63 & 0.66 & 0.58 \\ \hline 

\multicolumn{17}{c}{\textbf{GPT-4 Evaluation}} \\ \hline
\textbf{EC} & 0.72 & 0.79 & 0.71 & 0.62 & 0.76 & 0.85 & 0.76 & 0.63 & 0.76 & 0.82 & 0.73 & 0.72 & 0.81 & 0.80 & 0.80 & 0.84 \\
\textbf{CQ} & 0.61 & 0.68 & 0.60 & 0.52 & 0.64 & 0.78 & 0.61 & 0.49 & 0.61 & 0.68 & 0.59 & 0.54 & 0.66 & 0.70 & 0.63 & 0.63 \\
\textbf{FC} & 0.44 & 0.54 & 0.46 & 0.28 & 0.47 & 0.57 & 0.49 & 0.30 & 0.53 & 0.63 & 0.48 & 0.44 & 0.62 & 0.62 & 0.61 & 0.61 \\ \hline
\end{tabular}
\caption{Performance of Mixtral, Mixtral + RARR, Mixtral Revised and GPT-4 models for localized text generation on a \dataset subset using both human and GPT-4 evaluations. The subset consists 250 randomly sampled samples with 96, 83, and 71 samples from hyperlocal scores 1, 2, and 3 respectively.}
\label{table:models_250_subset}
\end{table*}

\begin{table*}[h]
    \centering
    \scriptsize
    \setlength{\tabcolsep}{4pt} 
    \begin{tabular}{lcp{3.33cm}cp{3.33cm}cp{3.33cm}}
    \hline
    \textbf{Reference Text} & \multicolumn{6}{p{12cm}}{The Free Spirits was an American band credited as the first jazz-rock group. The band also incorporated elements of pop and garage rock. Their first album Out of Sight and Sound was recorded in 1966 and released in 1967.} \\
    \textbf{Target Location} & \multicolumn{6}{p{12.5cm}}{Delhi}\\
    \textbf{Common Questions} & \multicolumn{6}{p{12.5cm}}{(i) Can you name a band credited as a jazz-rock band? (ii) Can you provide an example of a band which has released albums?}\\
     \textbf{True Target Entity} & \multicolumn{6}{p{12.5cm}}{Indian Ocean} \\
    \textbf{True Target Sentence} & \multicolumn{6}{p{12.5cm}}{The Indian Ocean is an Indian band credited as jazz-rock fusion group. The band also incorporated elements of folk and classical. Their first album Indian Ocean was recorded in 1992 and released in 1993.} \\
    \textbf{Mixtral} & \multicolumn{6}{p{12.5cm}}{The Mystic Moods, a Delhi-based band, was credited as one of the first to experiment with fusion of jazz and rock in India. They also incorporated elements of Indian classical music and pop. Their first album "A New Dawn" was recorded in 1972 and released in 1973.} \\ \hline
    \textbf{Evaluator} & \textbf{EC} & \centering \textbf{Reason} & \textbf{CQ} & \centering \textbf{Reason} & \textbf{FC} & \hspace{5em}\textbf{Reason} \\ \hline
    GPT-4 & 1 & The Mystic Moods is also a music band based in India, and shares a similar category with Indian Ocean, although it is not the exact same band. & [1, 1] & It mentions that 'The Mystic Moods,' a Delhi-based jazz-rock band and have released an album titled "A New Dawn". & 1 & The Mystic Moods is a Delhi-based jazz-rock band. It accurately describes their origin, genre, and album details, making it relevant and truthful in the context of Delhi.\\
    Human & 0 & \centering Entity does not exist. & [0, 0] & \centering EC = 0 & 0 &  \hspace{5em} EC = 0\\ \hline
    \end{tabular}
    \caption{An example to illustrate the limitation of GPT-4 as an evaluator for text localization by comparing it with human evaluation.}
    \label{table:eval_example}
\end{table*}
    
\subsection{Comparison of Models/Approaches for Localized Text Transfer}
 In Table \ref{table:models_250_subset}, we compare the performance of Mixtral, Mixtral + RARR, Mixtral Revised, and GPT-4 on a subset of 250 randomly chosen samples from \dataset using human evaluators.%
 \footnote{We restrict this evaluation to a 250-sample subset due to annotation costs. However, we point to the Mixtral and GPT-4 scores in Tables~\ref{table:mixtral_vs_gpt4} and~\ref{table:models_250_subset} which are very similar, thus affirming that the 250-sample subset of \dataset is representative of the full set.}
 Attribution using factual evidence helps Mixtral + RARR in improving Mixtral generations, especially in the CQ and FC metrics, where the scores improve by $9\%$ and $13\%$, respectively. However, the length of text obtained by RARR attribution is usually more than the original length of the text, and it fails to preserve the style.
 \looseness=-1

Mixtral Revised utilizes factual evidence similar to RARR but regenerates the text instead of editing it. Including factual evidence in the prompt enhances the Mixtral outputs and results in improvements in both FC and CQ. The approach focuses mainly on revising the factual correctness of the text while largely retaining the entity present in it. However, we still see an enhancement in EC as factual evidence provides a richer context for the effective localization of the entity. While both Mixtral Revised and Mixtral + RARR use evidence, the former re-generates the text and the latter edits the text by retaining the entity. Re-generation helps in obtaining a factually correct entity. GPT-4 surpasses all the Mixtral models due to its extensive training and diverse world knowledge. With increasing hyperlocal scores, even with GPT-4, performance degrades. Nonetheless, the revision step in Mixtral Revised significantly improves the scores across all metrics, particularly for regions with a hyperlocal score of 3. 

In Table \ref{table:models_250_subset}, we also analyze the capability of GPT-4 as an evaluator for the task of localized text transfer. Compared to humans, GPT-4 shows a $0.10-0.15$ increase across all the metrics and models, due to the presence of false positives. Table~\ref{table:eval_example} illustrates this limitation using an example. Mixtral hallucinates and returns the entity "Mystic Moods". GPT-4 incorrectly claims it is a factually correct localization and assigns a score of 1 for all the metrics. The comparison clearly shows that GPT-4 is not a reliable evaluator for absolute numbers. However, we observe similar overall trends in both human and GPT-4 evaluations. This shows that GPT-4 could be used as an LLM evaluator for localized text transfer to study the trends across models. Table \ref{table:subset_example} shows a detailed example for all the models discussed.
\looseness=-1

\begin{table*}[h]
    \centering
    \scriptsize
    \begin{tabular}{cp{10.5cm}ccc}
    \hline
    \textbf{Reference Text} & \multicolumn{4}{p{13cm}}{Avenged Sevenfold is an American heavy metal band from Huntington Beach, California, formed in 1999. The band's current lineup consists of vocalist M. Shadows.} \\
    \textbf{Target Location} & \multicolumn{4}{l}{Kolkata} \\
    \textbf{Common Questions} & \multicolumn{4}{l}{\begin{tabular}[c]{@{}l@{}}(i) Name a music band which was formed many years ago? (ii) Which band's current lineup consists of a vocalist? \vspace{3pt} \end{tabular}} \\ \hline 
    \textbf{Model} & \centering \textbf{Text Generated} & \textbf{EC} & \textbf{CQ} & \textbf{FC}\\ \hline
    Mixtral & Cactus is a popular Indian heavy metal band from Kolkata, formed in \textcolor{red}{2006}. The band's current lineup consists of vocalist \textcolor{red}{Bipul Das}, lead guitarist \textcolor{red}{Sidharth Roy}, rhythm guitarist \textcolor{red}{Swarnendu Biswas}, and drummer \textcolor{red}{Shayan Ghosh}. & 1 & {[}0, 0{]} & 0\\
    Mixtral + RARR & Cactus is a popular Indian heavy metal band from Kolkata, formed in \textcolor{green}{1992}. The current lineup consists of vocalist \textcolor{red}{Bipul Das}, lead guitarist \textcolor{red}{Sidharth Roy}, rhythm guitarist \textcolor{red}{Swarnendu Biswas}, and drummer \textcolor{red}{Shayan Ghosh}. \ul{The band has named Deep Purple, Pink Floyd, Led Zeppelin, Iron Maiden and many others as their inspiration. Cactus is one of the first professional Bangla rock bands, with their professional debut occurring on 6th March 1993 at Aban Mahal (CLT), Kolkata}. & 1 & {[}1, 0{]} & 0 \\
    Mixtral Revised & Cactus is a renowned Kolkata heavy metal band, formed in \textcolor{green}{1992}. The band's current lineup includes vocalists \textcolor{green}{Sidhartha Sankar "Sidhu" Ray and Abhijit "Pata" Barman}, guitarists \textcolor{green}{Boidurjyo Chowdhury and Samrat Banerjee}, \textcolor{green}{keyboardist Sayantan Chatterjee, bassist Proshanto Mahato,} and drummer \textcolor{green}{Arnab Tabla Dasgupta.} & 1 & {[}1, 1{]} & 1 \\
    GPT-4 & Fossils is an Indian rock band from Kolkata, West Bengal, formed in 1998. The band's current lineup consists of vocalist Rupam Islam. & 1 & {[}1, 1{]} & 1 \\ \hline
    \end{tabular}
    \begin{flushleft}
        \scriptsize{\hspace{1em}\textcolor{red}{Red} represents the error in the text, \textcolor{green}{green}} represents the correct edits and \ul{underline} represents the extra text generated.\\
    \end{flushleft}
    \caption{An example to illustrate the various text localization approaches: Mixtral, Mixtral + RARR, Mixtral Revised and GPT-4.}
    \label{table:subset_example}
\end{table*}

\subsection{\dataset as a Benchmark for Localized Question Answering}

\begin{table}[h]
    \centering
    \footnotesize
    \begin{tabular}{lcccc}
    \hline
        \textbf{Mixtral} & \textbf{Overall} &\textbf{1} & \textbf{2} & \textbf{3} \\ \hline
        \textbf{\# Samples} & 250 & 96 & 83 & 71 \\
        \textbf{\# Questions} & 447 & 168 & 145 & 134 \\ \hline
        \multicolumn{5}{c}{\textbf{Human Evaluation}} \\ \hline
        \multicolumn{1}{c}{\textbf{EC}} & 0.64 & 0.81 & 0.63 & 0.45 \\
        \multicolumn{1}{c}{\textbf{CQ}} & 0.63 & 0.79 & 0.60 & 0.45 \\
        \multicolumn{1}{c}{\textbf{FC}} & 0.59 & 0.77 & 0.58 & 0.37 \\ \hline
        \multicolumn{5}{c}{\textbf{GPT-4 Evaluation}} \\ \hline
        \multicolumn{1}{c}{\textbf{EC}} & 0.77 & 0.83 & 0.77 & 0.69 \\
        \multicolumn{1}{c}{\textbf{CQ}} & 0.71 & 0.79 & 0.70 & 0.62 \\
        \multicolumn{1}{c}{\textbf{FC}} & 0.61 & 0.74 & 0.53 & 0.52 \\ \hline
    \end{tabular}
    \caption{\dataset Dataset Benchmark for Localized Text Generation using Questions}
    \label{table:mixtral_using_cq}
    \vspace{-0.55cm}
\end{table}

\dataset can also be used as a benchmark to evaluate localized question answering. Given a target location and a question, the model has to generate text that answers the question while being correctly localized to the given target location. To aid this task, we also provide the reference location and the reference text as an example to guide localization and the style of generation.

Table \ref{table:mixtral_using_cq} shows the performance of promoting Mixtral on this benchmark task. Mixtral obtains accuracies of 64\%, 63\%, and 59\% on the EC, CQ, and FC metrics, respectively. Consistent with our previous observations, the model encounters challenges in efficient localization as hyperlocal scores increase. Some examples of Mixtral generations are shown in Table \ref{table:examples_mixtral_using_cq}. In Table \ref{table:mixtral_using_cq}, we also discuss the performance of GPT-4 as an evaluator for this benchmark task. GPT-4 nearly matches human evaluation when targeting India as a whole (hyperlocal score = 1), but highly overestimates scores for regions with hyperlocal scores of 2 and 3. The overall trends of human evaluation are maintained by GPT-4. We also show this comparison on the full \dataset dataset for the Mixtral model in Table ~\ref{table:mixtral_human_gpt4_1100}. 
\looseness=-1

\begin{table*}[h]
\centering
\scriptsize
\begin{tabular}{p{5.25cm}cp{2cm}p{2.75cm}ccc}
\hline
\centering \textbf{Reference Text} & \centering \textbf{Target Location} & \centering \textbf{Question} & \centering \textbf{Mixtral} & \textbf{EC} & \textbf{CQ} & \textbf{FC} \\ \hline
Morgan Freeman is an American actor, producer, and narrator known for his distinctive, deep voice. Throughout a career spanning five decades and multiple film genres, he has received numerous accolades, including an Academy Award, a Golden Globe Award, and a Screen Actors Guild Award as well as a nomination for a Tony Award. & Thanjavur & Name an actor who has worked for more than two decades. & Kamal Haasan is a renowned Indian actor who has been active in the film industry for over four decades, featuring in more than 200 films across various languages. & 0 & 0 & 0 \\ 
Duronto TV is a Bangladeshi Bengali-language satellite and cable children's television channel, and the first of its kind in the country. It was initially licensed in 2013 as "Renaissance TV" but was later renamed to its current name. & Mumbai & Can you name a children's television channel that is the first of its kind in a country? & Nickelodeon Sonic is a children's television channel in India, and it was the first of its kind when it launched in 2001. & 1 & 1 & 0 \\ \hline
\end{tabular}
\caption{Examples of Mixtral generations for the benchmark task of localized question answering.}
\label{table:examples_mixtral_using_cq}
\vspace{-0.2cm}
\end{table*}

\section{Discussion}
\paragraph{Localization.} The \dataset dataset caters to a very specific style of localization involving factual transfer. Localization can be much broader in scope extending to different vocabulary choices for measurements (miles vs. meters), daily objects (lift vs. elevator), food (cookie vs. biscuit), etc., depending on the target location. We elaborate on this further in Section~\ref{sec:limitations} and we intend to develop \dataset further to include more forms of localization.

\paragraph{GPT-4 evaluations.} From Table~\ref{table:models_250_subset}, we observe that human and GPT-4 evaluations are most similar for GPT-4 generations. For all other model generations, GPT-4 gives inflated scores for all metrics (particularly EC) compared to the human evaluations. But, the trends in GPT-4 evaluations across models for both EC and FC mimic the trends observed in human evaluations. (This is not as clear for the CQ metric.) This suggests that one could use GPT-4 evaluations (instead of very expensive human evaluations) to observe the trends in scores across multiple models to assess which model performs the best (or worst). We could enhance the GPT-4 evaluation with retrieval-augmented generation (RAG) techniques to improve its factuality assessments. We leave such enhancements for future work.

\section{Related Work}

\paragraph{Factual Correction, Transfer and Localization.} Improving factual accuracy of LM generations is a very important problem that has gathered recent interest. Evidence integration, LLM post-editing modules, Rank-One Model Editing (ROME) are some of the recent techniques used to correct factual errors but they all struggle with consistency, specificity and generalizability \citep{thorne2021evidencebased, cao2021factual, meng2023locating}. Evaluating factual accuracy is another important problem. FActScore~\citep{min2023factscore} is a fine-grained measure that decomposes a generation into multiple atomic facts and computes the fraction of facts supported by a knowledge source. This has also been extended to multilingual models~\citep{shafayat2024multifact}. However, all such measures are prone to biases across language and regions~\citep{mirza2024globalliar}. We empirically demonstrate such a regional bias using our \dataset dataset. 

In factual transfer, we also want the text style and intent of the reference text to be preserved as in standard text style transfer tasks~\citep{jin2021deep}. ModQGA is a framework that transfers facts without altering style~\citep{balepur2023text}. Techniques like inverse prompting~\citep{Zou2021ControllableGF} have been used to improve the generation quality of LLMs for factual transfer.
However LLMs struggle with self-correction, indicating limitations in such intrinsic mechanisms \citep{huang2024large}. The RARR system improves reliability and attribution by correcting unsupported content using external evidence \citep{gao2023rarr}. Hence, we adopt RARR as one of our approaches to test \dataset.


\paragraph{Cultural Adaptability and Diversity.} 
LLMs tend to be geographically biased on various dimensions such as culture, race, language, politics due to its training being dominated by Western/English-centric datasets~\citep{manvi2024large}. To address this challenge, CultureLLM uses semantic data augmentation to better represent multiple cultures \citep{li2024culturellm, li2024culturegen, li2024culturepark}. Another recent study shows that LLMs when evaluated on NORMAD dataset, struggle with cultural reasoning across different contexts, showing better adaptability to English-centric cultures compared to those from the Global South~\citep{rao2024normad}. In our work, we focus on an arguably simpler task of factual transfer across geographical regions for which there is no existing benchmark. 

\section{Conclusion}
This work introduces a new evaluation benchmark \dataset to test the localization and factual transfer capabilities of LLMs. We attempt to localize factual statements from across the globe to multiple target locations within India spanning different levels of hyperlocality. We establish various baselines (Mixtral, GPT-4, etc.) and multiple benchmark tasks for the different models. We find that GPT-4 struggles with localization at higher levels of hyperlocality (i.e., when localizing to Indian cities), so much so that it cannot be reliably used as an automatic evaluator. We hope \dataset helps the research community in designing improved localization and factual transfer techniques.

\clearpage
\section*{Limitations}
\label{sec:limitations}
The \dataset dataset is not without its limitations. A few of them are detailed below:
\begin{itemize}
    \item GPT-4 is not good at identifying hyperlocal entities and facts about them. Hence, it cannot be used to reliably evaluate whether or not the localization produced is correct. Thus, there is still a need for human evaluators to check whether the localization produced is correct or not. A possible remedy to this is to add multiple possible target entities and facts about them corresponding to each reference entity and the target location they are being localized for. This is something that we plan to eventually add to our dataset in the near future. We hope it will help eventually mitigate the need for human evaluators to check for correctness.
    \item There can be several correct target entities localized to a target location which we refer to as high cardinality. High cardinality can make it hard to make the resulting evaluations precise, especially since some entities can be added in the future with respect to localization. 
    \item This dataset consists only of factual data. However,  localization can take place with respect to actions as well. For example, suppose we are localizing a conversation between a human and a shopkeeper about a special dinner. In the west, this typically would include conversations about buying steaks, lobsters etc. while in India, the conversation would likely be more about buying spices, rice and chicken. This is a broader style of localization that we intend to explore further as future work.
    \item The dataset is designed for localization from different locations in the world to India only. In order to perform localization to regions other than in India, we will need additional annotations. This is also reserved for a future release.
    \item \dataset is entirely in English and does not contain any multilingual localizations. It is possible to use simple translation models to translate the data but it is not robust. This is a significant extension that we also intend to explore as future work.
\end{itemize}



\bibliography{latex/custom}

\appendix

\clearpage
\section{Appendix}
\renewcommand{\thefigure}{A\arabic{figure}}
\renewcommand{\thetable}{A\arabic{table}}
\setcounter{figure}{0}
\setcounter{table}{0}
\label{sec:appendix}

    \subsection{Annotation Process and Guidelines}
    \label{sec:annotation}
    The \dataset dataset was annotated by humans at various stages of its generation. The annotation was performed by an annotation company in India. The annotators were from diverse locations, occupations, age groups (21-40 yrs), and gender. The following guidelines were provided to the human annotators.
    \subsubsection{Generation of Entities}
        \begin{asparaitem}
            \item Entities should cover a diverse set of 99 categories. Examples of categories: Politician, Music Band, Historical Monument, Airline, Web Series, etc.
            \item On average 10 entity-pairs under each category. Note: Reference entity can be repeated, but do not repeat target entity.
            \item Ensure the target entity is sufficiently similar to the reference entity selected. For example, refer row 1 of Table \ref{table:editing_entity}.
            \item Ensure the new entities are spread over India and have different hyperlocal scores. For example, refer row 2-4 of Table \ref{table:editing_entity}.
            \item The reference entities of the dataset should be spread across different countries, with 60\% from the US/Europe and the remaining 40\% from other parts of the world.
        \end{asparaitem}
    
        \begin{table*}[h]
            \centering
            \scriptsize
            \begin{tabular}{|m{1.45cm}|m{2.25cm}|m{2.4cm}|m{2.25cm}|m{2.75cm}|m{1.75cm}|}
            \hline
            \textbf{Category} & \textbf{Reference Location} & \textbf{Reference Entity} & \textbf{Target Location} & \textbf{Target Entity} & \textbf{Hyperlocal Score}\\ \hline
            \label{r1} Singer	& US &	Taylor Swift	& India &	\text{\st{Neha Kakkar} Ravi Shankar} & 1\\ \hline
            Educational Institution	& Australia	& The University of Melbourne &	India	& Indian Institute of Technology, Bombay	&	1 \\ \hline
            Educational Institution	& Florida &	University of Central Florida	& Kerala	& Central University of Kerala & 2\\ \hline
            Educational Institution	& Miami &	University of Miami	& Tiruchirappali & Bharathidasan University & 3\\ \hline
            \end{tabular}
            \begin{flushleft}
            \footnotesize{\hspace{1em}\st{Text striked out} is the incorrect entity.}\\
            \end{flushleft}
            \caption{Example to illustrate how to create correct entity pairs for \dataset dataset.}
            \label{table:editing_entity}
        \end{table*}

    \subsubsection{Correction of Target Sentences}
        \begin{asparaitem}
            \item Check if the target sentences are factually correct and localized correctly.
            \item Altering multiple elements within the target sentence might be necessary to guarantee factual accuracy within the specific domain.
            \item Check for fluency, grammar, and vocabulary accuracy in the sentences while eliminating unnecessary symbols or words.
            \item Align the structure of the target sentence with that of the reference sentence. Remove or add any additional or missing content/information present in the reference sentence. For example, refer to Table \ref{table:editing_text}
        \end{asparaitem}
        
        \begin{table*}[h!]
            \centering
            \scriptsize
            \begin{tabular}{|m{1.5cm}|m{1.5cm}|m{1.5cm}|m{1.35cm}|m{1cm}|m{3cm}|m{3cm}|}
            \hline
            \textbf{Category} & \textbf{Reference Location} & \textbf{Reference Entity} & \textbf{Target Location} & \textbf{Target Entity}  & \textbf{Reference sentence} & \textbf{Target sentence} \\ \hline
            Automotive company	& US & Ford Motor & India & Tata Motors & \textcolor{red}{Ford Motor Company} is an \textcolor{blue}{American} multinational automobile manufacturer headquartered in \textcolor{orange}{Dearborn, Michigan, United States}. It was founded by \textcolor{teal}{Henry Ford} and incorporated on \textcolor{violet}{June 16, 1903}. & \textcolor{red}{Tata Motors Limited} is an \textcolor{blue}{Indian} multinational automotive \st{manufacturing company.} \textbf{[manufacturer headquartered in \textcolor{orange}{Mumbai, Maharashtra, India}]}. It was founded by \textcolor{teal}{J. R. D. Tata} and incorporated on \textcolor{violet}{September 1, 1945}. \st{The company sells passenger cars, trucks, vans, coaches, buses, sports cars, construction equipment and military vehicles under the Tata brand. Tata Motors is the largest automobile manufacturer in India with a revenue of over 470 billion Indian rupees.} \\ \hline
            \end{tabular}
            \begin{flushleft}
            \footnotesize{\textbf{[Text in square brackets]} is the additional content added and \st{Text striked out} is the additional content that has to be removed.}\\
            \end{flushleft}
            \caption{An example to illustrate the annotation process for the target sentence generated for the \dataset dataset.}
            \label{table:editing_text}
        \end{table*}

    \subsubsection{Common questions}
        \begin{asparaitem}
            \item It should be generated based on the common description of the entities in the pairs of text provided.
            \item It should be of the type such that it can be asked in any target location and still be valid.
            \item It should be free from specific details such as locations, timings, or unique identifiers connected to either event.
            \item Remove or correct any incorrect questions present. There should be a minimum of one correct common question for each sample sentence pair. Add more questions if needed.
        \end{asparaitem}
        For common question correction, refer to the example in Table \ref{table:editing_CQ}.
        
        \begin{table*}[h!]
            \centering
            \scriptsize
            \begin{tabular}{|m{4.9cm}|m{4.9cm}|m{4.9cm}|}
                \hline
                \textbf{Reference sentence} & \textbf{Target sentence} & \textbf{Common questions}\\ \hline
                Rishi Sunak is a British politician who has served as Prime Minister of the United Kingdom and Leader of the Conservative Party since 2022. & Narendra Modi is an Indian politician who has served as Prime Minister of India and President of the Bharatiya Janata Party since 2014. & (i) \textcolor{blue}{Can you name a current Prime Minister?} \newline (ii) \textcolor{blue}{Who is a well-known politician serving as the head of a major political party?}\\ \hline
                
                Poshmark is a social commerce marketplace where users can buy and sell new and secondhand fashion, home goods, and electronics. The platform has over 80 million users, with over 200M available listings. The company is headquartered in Redwood City, California, with offices in Canada, Australia, and India. & Meesho is a social commerce marketplace based in India where users can buy and sell new and secondhand fashion, home goods, and electronics. The platform has over 60 million users, with millions of available listings. The company is headquartered in Bengaluru, India, and operates independently. & (i) \textcolor{blue}{Name a social commerce marketplace ?} \newline (ii) \textcolor{blue}{Tell me about a company in the social commerce space?} \newline (iii) \textcolor{red}{Name a social commerce marketplace in California?} \newline (iv) \textcolor{red}{Who operates Poshmark as an independent subsidiary since January 2023?} \newline (v) \textcolor{red}{Where is Meesho headquartered, and do they have any connections to Naver Corporation or headquarters outside of India?} \\ \hline
                
                KLVE is a commercial radio station licensed to Los Angeles, California with a Spanish AC format. The station is owned by TelevisaUnivision, and is the flagship station for the Uforia Audio Network. &	Radio Mango 91.9 FM is a private radio station licensed to Kochi, Kerala with a Malayalam language format. The station is owned by the Malayala Manorama Group and serves as the flagship station for their radio network. & (i) \textcolor{red}{What kind of radio station is KLVE?} \newline (ii) \textcolor{red}{Who owns the radio station ""Radio Mango 91.9 FM""?} \newline (iii) \textcolor{blue}{Can you mention a radio station that is a flagship station for a network?} \\ \hline
        
                Mindhunter is an American psychological crime thriller television series created by Joe Penhall, which debuted in 2017, based on the 1995 true-crime book Mindhunter: Inside the FBI's Elite Serial Crime Unit by John E. Douglas and Mark Olshaker. & Kerala Crime Files is a Malayalam-language psychological crime drama web series directed Ahammed khabeer, which debuted in 2023. & (i) \textcolor{red}{Can you name an American psychological crime thriller television series that debuted in 2017?} \newline (ii) \textcolor{red}{Is there a television series based on a true-crime book that released recently?} \newline (iii) \textcolor{blue}{Mention a series focusing on criminal psychology.} \\ \hline
            \end{tabular}
            \begin{flushleft}
            \scriptsize{\textcolor{blue}{Blue} represents the correct questions and \textcolor{red}{Red} represents the incorrect questions.}\\
            \end{flushleft}
            \caption{Examples to illustrate the annotation process for the common question generated for the \dataset dataset.}
            \label{table:editing_CQ}
        \end{table*}

    \subsection{Human Evaluation Guidelines}
    The outputs generated by the models were evaluated by humans to assess Entity Correctness, Common Question Correctness, and Factual Correctness. The following guidelines were provided to the human annotators.
    \label{sec:human_eval_guidelines}
        \subsubsection{Entity Correctness (EC)}
        \begin{asparaitem}
            \item The entity detected from the sentence should be from the target location.
            \item Check if the entity is a correct localization of the reference entity provided.
            \item  If the entity is an exact match to the true target entity, please mention "Exact match" in the reason.
            \item Always provide a reason when the score is 0.
        \end{asparaitem}
        \subsubsection{Common Question Correctness (CQ)}
        \begin{asparaitem}
            \item Each sample will have multiple questions, evaluate each (sample, question) pair separately.
            \item For each sample, return the score as a list of 0's and 1's with the scores indexed at the question number.
            \item Common Question Correctness for all questions should be given a score of 0 if that sample's entity correctness (EC) is 0.
            \item Check if the sentence correctly answers the question for the "target location".
            \item Ensure factual correctness in these answers.
            \item Always provide a reason when the score is 0.
        \end{asparaitem}
        \subsubsection{Factual Correctness (FC)}
        \begin{asparaitem}
            \item Factual correctness should be given a score of 0, if that sample's entity correctness(EC) is 0.
            \item Assign a score of 1, if the sentence is fully factually correct, else assign a score of 0.
            \item If the sentence contains any information that lacks factual evidence online, assign a score of 0.
            \item Always provide a reason when the score is 0.
        \end{asparaitem}

    Refer to Table \ref{table:annotation_example}, for examples of human evaluation.

    \begin{table*}[h]
    \centering
    \scriptsize
    \renewcommand{\arraystretch}{2.5}
    \setlength{\tabcolsep}{4pt} 
    \begin{tabular}{m{0.7cm}m{1cm}m{1.cm}m{1cm}m{1cm}m{1cm}m{1.75cm}m{1.75cm}m{1.75cm}m{2cm}}
    \hline
    \textbf{Claim ID} & \textbf{Category} & \textbf{Reference Location} & \textbf{Reference Entity} & \textbf{Target Location} & \textbf{True Target Entity} & \textbf{Reference sentence} &  \textbf{True Target sentence} & \textbf{Questions} & \textbf{Model Generation}\\ \hline
    114	& Books & London & A Tale of Two Cities	& Cheruvathur & The House of Blue Mangoes &	A Tale of Two Cities is a historical novel published in 1859 by Charles Dickens, set in London and Paris before and during the French Revolution. & The House of Blue Mangoes is a historical novel published in 2002 by David Davidar, set in Chevathar during the historical events like caste conflicts, World War II, and India's struggle for independence. & (i) What is an example of a historical novel? \newline (ii) Can you name a novel that is set during the political change happening in the country? & "Kaalam Maarunnu Kaalangal" is a historical novel written in 1916 by C. V. Raman Pillai, set in Kilimanoor (Kerala) and Thiruvithamkoor (Travancore) during the late 18th and early 19th centuries.\\ 
    255	& Educational institution & US & Johns Hopkins University &	India &Indian Institute of Science	&	Johns Hopkins University is a private research university in Baltimore, Maryland. Founded in 1876, Johns Hopkins was the first U.S. university based on the European research institution model. &	The Indian Institute of Science is a public research university in Bangalore, Karnataka, India. Established in 1909, the Indian Institute of Science was the first to introduce a master's programs in engineering. & (i) Can you give an example of a research university? \newline (ii) Can you provide name of a university which introduced something new? & The Indian Institute of Science (IISc) is a public research university in Bangalore, Karnataka. Established in 1909, IISc was modeled after the European research institution and is one of the first institutions of its kind in India. \\ \hline
    \end{tabular}
    \begin{tabular}
    {m{0.7cm}m{1.5cm}m{2.6cm}m{1.5cm}m{3cm}m{1.5cm}m{3cm}}
    \multicolumn{7}{c}{\textbf{Evaluation}}\\ \hline
    \textbf{Claim ID} & \textbf{EC} & \textbf{Reason} & \textbf{FC} & \textbf{Reason} & \textbf{CQ} & \textbf{Reason} \\ \hline
    
    114	& 0	& wrong target location	& 0	& EC = 0 & [0, 0] & EC = 0\\ 
    255	& 1	& Exact match & 0	& some details like European are incorrect & [1, 0]	& (ii)The sentence does not answer the question.\\ \hline
    \end{tabular}
    \caption{Examples to illustrate human evaluation.}
    \label{table:annotation_example}
    \end{table*}

    \begin{figure*}[h!]
            \centering
            \begin{tcolorbox}[colback=white,colframe=black, boxrule=0.2mm, fontupper=\footnotesize, left=0.1mm, right=0.1mm, top=0.1mm, bottom=0.1mm]
            You are tasked with generating basic questions from common property or common description of the entities in pairs of sentences provided. The goal is to create 2 or more questions such that they can be asked in any location and still be valid. The questions should not have any entity or location mentioned in it. Example:\par \vspace{2pt}
            Given the following pair of sentences: \par
            (1) Poshmark is a social commerce marketplace where users can buy and sell new and secondhand fashion, home goods, and electronics. The platform has over 80 million users, with over 200M available listings. The company is headquartered in Redwood City, California, with offices in Canada, Australia, and India; \par
            (2) Meesho is a social commerce marketplace based in India where users can buy and sell new and secondhand fashion, home goods, and electronics. The platform has over 60 million users, with millions of available listings. The company is headquartered in Bengaluru, India, and operates independently. \par \vspace{2pt}
            The correct questions are: \par
            (i) Name a social commerce marketplace. \par
            (ii) Tell me about a company in the social commerce space. \par \vspace{2pt}
            The wrong questions are: \par
            (i) Name a social commerce marketplace in California. \par
            (ii) Who operates Poshmark as an independent subsidiary since January 2023? \par
            (iii) Where is Meesho headquartered, and do they have any connections to Naver Corporation or headquarters outside of India? \par \vspace{2pt}
            As shown in the examples, the correct questions should be free from specific details such as locations, timings, or unique identifiers connected to either event. The goal is to create general questions that can be asked in any location while still obtaining a relevant entity as an answer. Keep the questions simple. \par \vspace{2.5pt}
            Now generate only correct questions for the following pair: \par
            Sentence 1: \textbf{Rishi Sunak is a British politician who has served as Prime Minister of the United Kingdom and Leader of the Conservative Party since 2022.}\par
            Sentence 2: \textbf{Narendra Modi is an Indian politician who has served as Prime Minister of India, since 2014 and is a member of the Bharatiya Janata Party. }
            \end{tcolorbox}
            \caption{\centering Few-shot prompt for common question generation on Mixtral}
            \label{fig:prompt_cq}
        \end{figure*}

    \begin{figure*}[h!]
        \centering
        \begin{tcolorbox}[colback=white,colframe=black, boxrule=0.2mm, fontupper=\footnotesize, left=0.1mm, right=0.1mm, top=0.1mm, bottom=0.1mm]
          You are a localization assistant. Convert the reference entity sentence from English to the Indian domain by replacing the source entity with the target entity. Make the needed modifications in the sentence to make it factually correct for the target entity. Output answers in English using multi-entity localization. \par \vspace{2pt}
            Reference entity: \textbf{Rishi Sunak} \par
            Reference entity location: \textbf{UK} \par
            Reference entity sentence: \textbf{Rishi Sunak is a British politician who has served as Prime Minister of the United Kingdom and Leader of the Conservative Party since 2022.} \par
            Target entity: \textbf{Narendra Modi} \par
            Target entity location: \textbf{India}
        \end{tcolorbox}
        \caption{\centering Prompt for text localization on Mixtral}
        \label{fig:prompt_loc}
    \end{figure*}

    \subsection{Implementation Details}
   We ran our \emph{Mixtral-8x7b-instruct-v0.1.Q4\_K\_M} model experiments on a single NVIDIA DGX A100 GPU. A maximum sequence length of 32768 was used. For GPT-4 experiments we used the \emph{gpt-4-turbo} version of OpenAI.
    
    \subsection{Prompts used for Dataset Creation}
    The prompt used for text localization is given in Figure ~\ref{fig:prompt_loc}. And the prompt used for the generating of common questions from the reference and target text is given in Figure ~\ref{fig:prompt_cq}.
        
    \subsection{Category Distribution of \dataset dataset}
    
\begin{figure}[h!]
            \centering
            \includegraphics[width=\linewidth, height = 23cm]{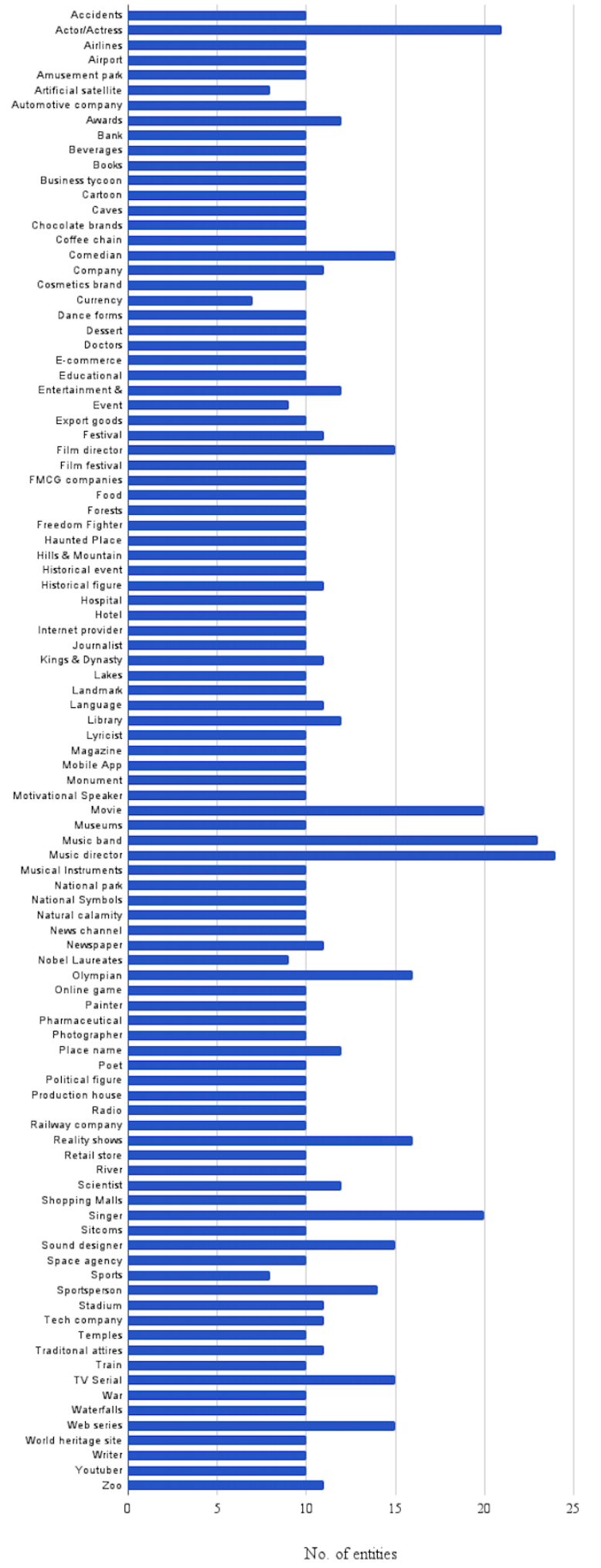}
            \caption{\centering \dataset dataset category distribution}
            \label{fig:count_category}
        \end{figure}

    The \dataset dataset contains 99 diverse categories like Movies, Accidents, Currency, Sports, etc. The number of entities under each category is uniformly distributed with an average of 10 entities in each category. Figure ~\ref{fig:count_category} shows the distribution of entities across the categories. As shown in Table ~\ref{table:categories}, categories can be grouped mainly into 10 clusters namely Entertainment, Professions, Buildings/Monuments/Companies, Food \& Lifestyle, Places \& Landmarks, Nature, Sports, Incidents, Finance \& Economy and Others. 

    \begin{table}[h!]
            \scriptsize
            \centering
            \setlength{\tabcolsep}{5pt} 
            \begin{tabular}{|p{1.4cm}|p{4.3cm}|p{0.75cm}|}
            \hline
            \textbf{Category Cluster} & \multicolumn{1}{|c|}{\textbf{Categories}} & \textbf{\# \newline Samples} \\ \hline
            Entertainment & Actor/Actress, TV Serial, Cartoon, Film Festival, Event, Magazine, Mobile App, Movie, Music Band, Radio, Sitcoms, Online Game, Web Series, News Channel, Newspaper, Production House, Awards, Books, Reality Shows, Dance Forms, Musical Instruments, Entertainment \& Sports Channel & 274 \\
            Professions & Business Tycoon, Comedian, Doctors, Film Director, Lyricist, Journalist, Motivational Speaker, Music Director, Poet, Writer, Singer, Scientist, Painter, Youtuber, Sound Designer, Photographer, Political Figure, Nobel Laureates & 220 \\ 
            Buildings/ Monuments/ Companies & Automotive Company, Company, Airlines, Educational Institution, FMCG Companies, Hospital, Hotel, Library, Temples, Pharmaceutical Companies, Airport, Tech Company, Space Agency, Monument, Railway Company, Museums, Internet Provider & 174 \\ 
            Others & Traditional Attires, Train, Language, Kings \& Dynasty, National Symbols, Artificial Satellite, Historical Figure, Festival, Freedom Fighter & 93 \\ 
            Food \& Lifestyle & Beverages, Chocolate Brands, Coffee Chain, Cosmetics Brand, Food, Dessert, Shopping Malls, Retail store, E-commerce company & 90 \\
            Places \& Landmarks & Landmark, Place Name, Haunted Place, Zoo, Amusement Park, National Park, World Heritage Site & 73 \\ 
            Nature & Caves, Forests, Hills \& Mountains, Lakes, River, Waterfalls & 60 \\ 
            Sports & Sports, Sportsperson, Olympian, Stadium & 49 \\ 
            Incidents & Historical Event, War, Accidents, Natural Calamity& 40\\ 
            Finance \& Economy & Bank, Currency, Export Goods & 27 \\ \hline
            \end{tabular}
            \caption{Category Clusters and Categories in \dataset dataset.}
            \label{table:categories}
        \end{table}

    \begin{table}[h!]
        \scriptsize
        \centering
        \setlength{\tabcolsep}{3pt} 
        \begin{tabular}{c|c|ccc|ccc}
        \hline
        \multicolumn{1}{c|}{\multirow{2}{*}{\textbf{Category Cluster}}} & \multicolumn{1}{c|}{\multirow{2}{*}{\textbf{\#}}} & \multicolumn{3}{c|}{\textbf{Mixtral}} & \multicolumn{3}{c}{\textbf{GPT-4}} \\ \cline{3-8}

        \multicolumn{1}{c|}{} & \multicolumn{1}{c|}{} & \multicolumn{1}{c}{\textbf{EC}} & \multicolumn{1}{c}{\textbf{FC}} & \multicolumn{1}{c|}{\textbf{CQ}} & \multicolumn{1}{c}{\textbf{EC}} & \multicolumn{1}{c}{\textbf{CQ}} & \multicolumn{1}{c}{\textbf{FC}} \\ \hline
        
        Entertainment & 274 & \multicolumn{1}{c}{0.57} & \multicolumn{1}{c}{0.27} & 0.44 & \multicolumn{1}{c}{0.75} & \multicolumn{1}{c}{0.57} & 0.56 \\
        
        Professions & 220 & \multicolumn{1}{c}{0.56} & \multicolumn{1}{c}{0.25} & 0.45 & \multicolumn{1}{c}{0.71} & \multicolumn{1}{c}{0.55} & 0.54 \\ 
        
        Building/Monument/Company & 174 & \multicolumn{1}{c}{0.80} & \multicolumn{1}{c}{0.59} & 0.68 & \multicolumn{1}{c}{0.88} & \multicolumn{1}{c}{0.76} & 0.71 \\ 
        
        Others & 93 & \multicolumn{1}{c}{0.71} & \multicolumn{1}{c}{0.52} & 0.56 & \multicolumn{1}{c}{0.88} & \multicolumn{1}{c}{0.75} & 0.79 \\ 
        
        Food \& Lifestyle & 90 & \multicolumn{1}{c}{0.67} & \multicolumn{1}{c}{0.35} & 0.51 & \multicolumn{1}{c}{0.83} & \multicolumn{1}{c}{0.70} & 0.65 \\
        
        Places \& Landmarks & 73 & \multicolumn{1}{c}{0.81} & \multicolumn{1}{c}{0.46} & 0.63 & \multicolumn{1}{c}{0.93} & \multicolumn{1}{c}{0.75} & 0.73 \\ 
        
        Nature & 60 & \multicolumn{1}{c}{0.75} & \multicolumn{1}{c}{0.42} & 0.58 & \multicolumn{1}{c}{0.83} & \multicolumn{1}{c}{0.65} & 0.58 \\ 
        
        Sports & 49 & \multicolumn{1}{c}{0.61} & \multicolumn{1}{c}{0.24} & 0.35 & \multicolumn{1}{c}{0.79} & \multicolumn{1}{c}{0.56} & 0.60 \\ 
        
        Incidents & 40 & \multicolumn{1}{c}{0.57} & \multicolumn{1}{c}{0.30} & 0.45 & \multicolumn{1}{c}{0.75} & \multicolumn{1}{c}{0.63} & 0.50 \\ 
        
        Finance \& Economy & 27 & \multicolumn{1}{c}{0.85} & \multicolumn{1}{c}{0.63} & 0.67 & \multicolumn{1}{c}{0.74} & \multicolumn{1}{c}{0.66} & 0.63 \\ \hline
        \end{tabular}
        \caption{Category-wise Performance Analysis of Mixtral and GPT-4 Generation}
        \label{table:category_analysis}
        \end{table}
    
    \subsection{Category-wise Performance Analysis of Models}
    In this section, we compare the performance of Mixtral and GPT-4 outputs across different categories. The \dataset has 99 unique categories and we have grouped them into 10 category clusters for our analysis. 
    
    Table~\ref{table:category_analysis} shows that the performance varies across categories `Professions', `Entertainment', and `Incidents' obtain the lowest scores by Mixtral and GPT-4 models due to the presence of diverse entities like Web Series, Movies, YouTubers, Motivational speakers, Accidents, etc. that have higher cardinality and lack of factual evidence. Both Mixtral and GPT-4 perform well in categories like `Buildings/Monuments/Companies', `Places \& Landmarks', and `Nature' due to the sufficient amounts of factual evidence available during training.

    \subsection{Prompt for localized text transfer} \label{sec: zero_shot_prompt}
        The prompt used for localized text transfer is given in Figure ~\ref{fig:zero_shot_prompt}. We use the same prompt for both Mixtral and GPT-4 models.
        \begin{figure*}[h]
            \centering
            \begin{tcolorbox}[colback=white,colframe=black, boxrule=0.2mm, fontupper=\footnotesize, left=0.1mm, right=0.1mm, top=0.1mm, bottom=0.1mm]
            You are a localization assistant. Convert the reference entity sentence from English to the Indian domain by localizing the reference sentence with a similar entity from the target location. Make the needed modifications in the sentence to make it factually correct for the target location. Output answers in English using multi-entity localization. Use the below format. \par \vspace{2pt}
            My reference sentence: <reference\_claim> \par
            Target location: <target\_location> \par
            Target sentence: <localized\_target\_sentence> \par
            Reason: <reason\_for\_the\_localization> \par \vspace{2pt}
            My reference sentence: \textbf{Rishi Sunak is a British politician who has served as Prime Minister of the United Kingdom and Leader of the Conservative Party since 2022.} \par
            Target location: \textbf{India} \par
            Target sentence: <fill\_your\_answer\_here> \par
            Reason: <fill\_your\_answer\_here> \par
            \end{tcolorbox}
            \caption{The prompt used for localized text transfer in Mixtral and GPT-4 models.}
            \label{fig:zero_shot_prompt}
        \end{figure*}

    \subsection{Prompt for localized question naswering}
        The rompt used for localized question answering is given in Figure ~\ref{fig:zero_shot_prompt_question}. We use the same prompt for both Mixtral and GPT-4 models.
        \begin{figure*}[h]
            \centering
            \begin{tcolorbox}[colback=white,colframe=black, boxrule=0.2mm, fontupper=\footnotesize, left=0.1mm, right=0.1mm, top=0.1mm, bottom=0.1mm]
            Given a question and a target location, generate a factually correct sentence such that it answers the given question using an entity from the target location. A reference location and sentence is given as an example. Output the answers in English. Use the below format. \par \vspace{2.5pt}
            Question: <question\_to\_be\_answered> \par
            Reference location: <example\_reference\_location> \par
            Reference sentence: <example\_reference\_sentence> \par
            Target location: <target\_location> \par
            Target sentence: <target\_sentence> \par
            Reason: <reason\_for\_the\_localization> \par \vspace{2.5pt}
            Question: \textbf{Can you name a current Prime Minister?} \par
            Reference location: \textbf{UK} \par
            Reference sentence: \textbf{Rishi Sunak is a British politician who has served as Prime Minister of the United Kingdom and Leader of the Conservative Party since 2022.} \par
            Target location: \textbf{India} \par
            Target sentence: <fill\_your\_answer\_here> \par
            Reason: <fill\_your\_answer\_here> \par
            \end{tcolorbox}
            \caption{The prompt used for localized question answering in Mixtral and GPT-4 models.}
            \label{fig:zero_shot_prompt_question}
        \end{figure*}  
    
    \subsection{Mixtral + RARR Prompts} 
    \label{sec:RARR}
        The prompts used in the Question Generation module, Evidence Retrieval module (to check whether the evidence agrees/disagrees with the text to be edited), and the Editor module are given in Figure ~\ref{fig:rarr_ques_gen}, ~\ref{fig:rarr_verification} and ~\ref{fig:rarr_editor} respectively.
        
        \begin{figure*}[h]
            \centering
            \begin{tcolorbox}[colback=white,colframe=black, boxrule=0.2mm, fontupper=\footnotesize, left=0.1mm, right=0.1mm, top=0.1mm, bottom=0.1mm]
            Given a target sentence corresponding to a specific target location, your task is to ask questions about the target entity. Each question should be specific to the target entity and should not contain pronouns such as 'he,' 'she,' 'it,' or 'they.' The questions should seek relevant information about the target entity, its attributes, actions, or associations with the target location. Additionally, the questions should be structured in a way that the answer contains the target entity and/or the target location. Avoid general questions like 'Who is he?' or 'Where does he live?' Instead, focus on extracting detailed insights about the target entity. Ensure that the questions are clear, concise, and relevant to the context of the target sentence. Questions should be able to interrogate the factual information in the claim. Do not generate irrelevant questions based on other entities that have no relation with the target entity or target locations. \par \vspace{2.5pt}
            For example: \par
            Target location: Delhi \par
            Target sentence: The India Gate is a war memorial made of sandstone located in the heart of New Delhi, India. It is named after the engineer Sir Edwin Lutyens, who designed and built the monument in 1931 to honor the Indian soldiers who died during World War I and the Third Anglo-Afghan War. \par
            The questions in the context of target sentence and target location are as follows: \par
            Q: What material is the India Gate made of? \par
            Q: In which city is the India Gate located?\par
            Q: Who is the engineer credited with designing and building the India Gate? \par
            Q: When was the India Gate constructed? \par \vspace{2.5pt}

            Target location: India \par
            Target sentence: Narendra Modi is an Indian politician who has served as Prime Minister of India and leader of the Bharatiya Janata Party since 2014. \par
            The questions in the context of target sentence and target location are as follows: \par
            Q: Who has served as the Prime Minister of India since 2014? \par
            Q: In which country does Narendra Modi hold the position of Prime Minister? \par
            Q: What is the name of the political party led by Narendra Modi, which is based in India? \par
            Q: Who has been the leader of the Bharatiya Janata Party in India since 2014? \par \vspace{2.5pt}
            
            Target location: India \par
            Target sentence:  Cricket is a bat-and-ball sport played between two teams of eleven players each, taking turns batting and fielding. The game occurs over the course of several overs, with each over consisting of six deliveries (pitches) generally made by a player on the fielding team, called the bowler, which a player on the batting team, called the batter, tries to hit with a bat. \par
            The questions in the context of target sentence and target location are as follows:
            Q: What sport is commonly played between two teams of eleven players each in India? \par
            Q: In India, what is the name of the player on the fielding team who delivers the ball to the batter? \par
            Q: What is the objective of the batter in the sport commonly played in India?
            Q: In India, what is the term for a single set of deliveries made by a bowler in the sport? \par
            Q: What sport, played in India, involves teams taking turns batting and fielding? \par \vspace{2.5pt}

            Target location: Telengana \par
            Target sentence: S. S. Rajamouli is an Indian director and screenwriter, known for his work in Telugu industry based in Telengana, India. He is considered one of the leading filmmakers in the Indian film industry, having directed some of the highest-grossing Indian films of all time. His most notable works include the "Telugu-language fantasy action film series", Baahubali and RRR which broke several box office records and gained international recognition. \par
            The questions in the context of target sentence and target location are as follows: \par
            Q: In which Indian state is S. S. Rajamouli primarily associated with for his filmmaking? \par
            Q: What are some of the notable works directed by S. S. Rajamouli in the Telugu film industry? \par
            Q: What Indian state is known for its flourishing Telugu film industry, where S. S. Rajamouli has made significant contributions? \par
            Q: Which Indian filmmaker is renowned for directing the "Telugu-language fantasy action film series" Baahubali and RRR? \par \vspace{2.5pt}
            
            Target location: Bengaluru \par
            Target sentence: "Flipkart" is an Indian e-commerce company headquartered in Bengaluru, Karnataka. Founded by Sachin Bansal and Binny Bansal in 2007, it started as an online bookstore before diversifying into a wide range of product categories, including electronics, fashion, and home goods. With its user-friendly interface, extensive product offerings, and competitive pricing, Flipkart has emerged as one of India's leading e-commerce platforms, revolutionizing the way millions of people shop online in the country. \par
            The questions in the context of target sentence and target location are as follows: \par
            Q: What is the name of the Indian e-commerce company headquartered in Bengaluru, Karnataka? \par
            Q: Who are the founders of Flipkart, the Indian e-commerce company based in Bengaluru? \par
            Q: In which Indian city is Flipkart headquartered? \par
            Q: What year was Flipkart founded by Sachin Bansal and Binny Bansal in Bengaluru? \par
            Q: How has Flipkart impacted the way millions of people shop online in India? \par \vspace{2.5pt}

            Target location: \textbf{West Bengal} \par
            Target sentence: \textbf{Prosenjit Chatterjee is a renowned Bengali actor, whose career has been marked by critical acclaim in his early life, followed by personal challenges and a resurgence in popularity and commercial success in his later years.} \par
            The questions in the context of target sentence and target location are as follows: \par
            \end{tcolorbox}
            \caption{Mixtral + RARR: The prompt used for generating questions from the sentence and target location for evidence retrieval.}
            \label{fig:rarr_ques_gen}
        \end{figure*}

        \begin{figure*}[h]
            \centering
            \begin{tcolorbox}[colback=white,colframe=black, boxrule=0.2mm, fontupper=\footnotesize, left=0.1mm, right=0.1mm, top=0.1mm, bottom=0.1mm]
            I will check some things you said. \par \vspace{2.5pt}
            1. You said: Your nose switches back and forth between nostrils. When you sleep, you switch about every 45 minutes. This is to prevent a buildup of mucus. It’s called the nasal cycle. \par
            2. I checked: How often do your nostrils switch? \par
            3. I found this article: Although we don’t usually notice it, during the nasal cycle one nostril becomes congested and thus contributes less to airflow, while the other becomes decongested. On average, the congestion pattern switches about every 2 hours, according to a small 2016 study published in the journal PLOS One. \par
            4. Reasoning: The article said the nose’s switching time is about every 2 hours, and you said the nose's switching time is about every 45 minutes. \par
            5. Therefore: This disagrees with what you said. \par \vspace{2.5pt}
            
            1. You said: The Little House books were written by Laura Ingalls Wilder. The books were published by HarperCollins. \par
            2. I checked: Who published the Little House books? \par
            3. I found this article: These are the books that started it all -- the stories that captured the hearts and imaginations of children and young adults worldwide. Written by Laura Ingalls Wilder and published by HarperCollins, these beloved books remain a favorite to this day. \par
            4. Reasoning: The article said the Little House books were published by HarperCollins and you said the books were published by HarperCollins. \par
            5. Therefore: This agrees with what you said. \par \vspace{2.5pt}
            
            1. You said: Real Chance of Love was an American reality TV show. Season 2 of the show was won by Cali, who chose to be with Chance. \par
            2. I checked: Who won season 2 of Real Chance of Love? \par
            3. I found this article: Real Chance of Love 2: Back in the Saddle is the second season of the VH1 reality television dating series Real Chance of Love. Ahmad Givens (Real) and Kamal Givens (Chance), former contestants on I Love New York are the central figures. \par
            4. Reasoning: The article doesn't answer the question and you said that Cali won season 2 of Real Chance of Love. \par
            5. Therefore: This is irrelevant to what you said. \par \vspace{2.5pt}
            
            1. You said: The Havel-Hakimi algorithm is an algorithm for converting the adjacency matrix of a graph into its adjacency list. It is named after Vaclav Havel and Samih Hakimi. \par
            2. I checked: What is the Havel-Hakimi algorithm? \par
            3. I found this article: The Havel-Hakimi algorithm constructs a special solution if a simple graph for the given degree sequence exists, or proves that one cannot find a positive answer. This construction is based on a recursive algorithm. The algorithm was published by Havel (1955), and later by Hakimi (1962). \par
            4. Reasoning: The article said the Havel-Hakimi algorithm is for constructing a special solution if a simple graph for the given degree sequence exists and you said the Havel-Hakimi algorithm is for converting the adjacency matrix of a graph. \par
            5. Therefore: This disagrees with what you said. \par \vspace{2.5pt}
            
            1. You said: "Time of My Life" is a song by American singer-songwriter Bill Medley from the soundtrack of the 1987 film Dirty Dancing. The song was produced by Michael Lloyd. \par
            2. I checked: Who was the producer of "(I’ve Had) The Time of My Life"? \par
            3. I found this article: On September 8, 2010, the original demo of this song, along with a remix by producer Michael Lloyd , was released as digital files in an effort to raise money for the Patrick Swayze Pancreas Cancer Resarch Foundation at Stanford University. \par
            4. Reasoning: The article said that a demo was produced by Michael Lloyd and you said "Time of My Life" was produced by Michael Lloyd. \par
            5. Therefore: This agrees with what you said. \par \vspace{2.5pt}
            
            1. You said: Tiger Woods is the only player who has won the most green jackets. He has won four times. The Green Jacket is one of the most coveted prizes in all of golf. \par
            2. I checked: What is the Green Jacket in golf? \par
            3. I found this article: The green jacket is a classic, three-button, single-breasted and single-vent, featuring the Augusta National Golf Club logo on the left chest pocket. The logo also appears on the brass buttons. \par
            4. Reasoning: The article said the Green Jacket is a classic three-button single-breasted and single-vent and you said the Green Jacket is one of the most coveted prizes in all of golf. \par
            5. Therefore: This is irrelevant to what you said. \par  \vspace{2.5pt}
            
            1. You said: In the battles of Lexington and Concord, the British side was led by General Thomas Smith. \par
            2. I checked: Who led the British side in the battle of Lexington and Concord? \par
            3. I found this article: Interesting Facts about the Battles of Lexington and Concord. The British were led by Lieutenant Colonel Francis Smith. There were 700 British regulars. \par
            4. Reasoning: The article said the British side was led by Lieutenant Colonel Francis Smith and you said the British side was led by General Thomas Smith. \par
            5. Therefore: This disagrees with what you said. \par  \vspace{2.5pt}
            
            1. You said: \textbf{Prosenjit Chatterjee is a renowned Bengali actor, whose career has been marked by critical acclaim in his early life, followed by personal challenges and a resurgence in popularity and commercial success in his later years.} \par
            2. I checked: \textbf{West Bengal: What type of recognition has marked Prosenjit Chatterjee's early life in his film career?} \par
            3. I found this article: \textbf{June 4, 2023 National recognition and accolades did not lure Prosenjit away from West Bengal's entertainment industry. He began this year with a stellar performance in Kaushik Ganguly's period thriller, Kaberi Antardhan, shot against the backdrop of the Naxalite movement and the Emergency.} \par
            4. Reasoning: \par
            \end{tcolorbox}
            \caption{Mixtral + RARR: The prompt used by RARR \cite{gao2022rarr} for checking the agreement of the retrieved evidence for editing.}
            \label{fig:rarr_verification}
        \end{figure*}

        \begin{figure*}[h]
            \centering
            \begin{tcolorbox}[colback=white,colframe=black, boxrule=0.2mm, fontupper=\footnotesize, left=0.1mm, right=0.1mm, top=0.1mm, bottom=0.1mm]
            This task involves processing a claim by attributing it based on a set of evidences. The aim is to refine the initial claim into an attributed claim that incorporates insights from all provided evidences. \par
            Instructions:\par
            1. Identify the main entity discussed in the provided claim. Carefully review all associated evidences. Note that the evidences may or may not be relevant to the main entity of the claim. \par
            2. Determine the relevance of each piece of evidence to the main entity in the claim. Synthesize the factual information from relevant evidences to assess how they support, refute, or modify the initial claim.  \par
            3. Generate an attributed claim that effectively integrates the initial claim with the relevant evidences, ensuring that the main entity of the claim remains unchanged, especially in the context of any irrelevant evidence. \par
            4. Do not include unnecessary evidence sentences in the modified claim which were not present in the original claim. You are required to check only the factual correctness of the claim without adding extra information to the claim.  \par \vspace{2.5pt}
            
            Example: \par  \vspace{2.5pt}
            
            Claim: Tata Motors is an Indian multinational automobile manufacturing company headquartered in Mumbai, Maharashtra, India. It was established in 1954. \par
            Evidences: \par
            1. Mahindra \& Mahindra Limited (M\&M) is an Indian multinational automotive manufacturing corporation headquartered in Mumbai. It was established in 1945 as Mahindra \& Mohammed and later renamed Mahindra \& Mahindra. \par
            2. Tata Motors was founded in 1945, as a locomotive manufacturer. Tata Group entered the commercial vehicle sector in 1954 after forming a joint venture with Daimler-Benz of Germany in which Tata developed a manufacturing facility in Jamshedpur for Daimler lorries. \par
            Attributed Claim: Tata Motors is an Indian multinational automobile manufacturing company headquartered in Mumbai, Maharashtra, India. It was established in 1945. \par  \vspace{2.5pt}
            
            Claim: Feluda is a detective novel written by renowned Bengali actor Sandip Ray, first published in West Bengal in 1965 by Ananda Publishers. The book has been adapted into a film and several television series. \par
            Evidences: \par
            1. Feluda is an Indian-Bengali detective media franchise created by Indian-Bengali film director and writer Satyajit Ray, featuring the character, Feluda. \par
            2. In 1965, at the age of 44, soon after the release of his landmark film Charulata, Satyajit Ray wrote the first draft of a short story, which featured a young boy, barely into his teens, describing the superlative analytical and detection powers of his older cousin brother." \par
            Attributed Claim: "Feluda is a detective novel written by renowned Bengali author Satyajit Ray, first published in West Bengal in 1965 by Ananda Publishers. The book has been adapted into a film and several television series. \par  \vspace{2.5pt}

            Claim: Leonardo DiCaprio won his first Oscar for Best Actor for his role in the film 'Titanic' in 1996. \par
            Evidences: \par
            1. Leonardo DiCaprio has been nominated for the Best Actor Oscar multiple times, beginning with his role in 'What's Eating Gilbert Grape' in 1993.\par
            2. DiCaprio's performance in 'The Revenant' was universally acclaimed, and he won the Academy Award for Best Actor in 2016, which was his first Oscar win. \par
            3. Leonardo DiCaprio is an active environmentalist who has donated millions to conservation efforts. \par
            Attributed Claim: Leonardo DiCaprio won his first Oscar for Best Actor for his role in 'The Revenant' in 2016, after several nominations for other films including his first for 'What's Eating Gilbert Grape.'  \par  \vspace{2.5pt}
            
            Claim: Avengers: Endgame was released worldwide in April 2018 and became the highest-grossing film of all time by surpassing 'Titanic'. \par
            Evidences: \par
            1. Avengers: Endgame was released in April 2019. It quickly garnered acclaim for its dramatic conclusion of the Infinity Saga." \par
            2. In July 2019, 'Avengers: Endgame' surpassed 'Avatar' to become the highest-grossing film ever, a record it held until 'Avatar' reclaimed the title after a re-release." \par
            3. The soundtrack for 'Avengers: Endgame' was composed by Alan Silvestri, who also composed music for 'Back to the Future.'" \par
            Attributed Claim: Avengers: Endgame was released worldwide in April 2019 and became the highest-grossing film of all time by surpassing 'Avatar' in July of that year, although 'Avatar' later reclaimed the top spot. \par  \vspace{3.5pt}

            For this claim and evidences, generate the attributed claim as instructed. \par  \vspace{2.5pt}
            Claim: \textbf{Prosenjit Chatterjee is a renowned Bengali actor, whose career has been marked by critical acclaim in his early life, followed by personal challenges and a resurgence in popularity and commercial success in his later years.} \par
            Evidences: \par
            \textbf{1. June 4, 2023 National recognition and accolades did not lure Prosenjit away from West Bengal's entertainment industry. He began this year with a stellar performance in Kaushik Ganguly's period thriller, Kaberi Antardhan, shot against the backdrop of the Naxalite movement and the Emergency.} \par
            Attributed Claim: \par
            \end{tcolorbox}
            \caption{Mixtral + RARR: The prompt used for the non-sequential editing of the text.}
            \label{fig:rarr_editor}
        \end{figure*}

    \subsection{Mixtral Revised Prompts} \label{sec:mixtral_revised}
        The prompt used for verifying the relevance of the evidence for the target context is given in Figure ~\ref{fig:revised_verify}. The text re-generation prompt of the Mixtral Revised model is given in Figure ~\ref{fig:revised_gen}.
        \begin{figure*}[h]
            \centering
            \begin{tcolorbox}[colback=white,colframe=black, boxrule=0.2mm, fontupper=\footnotesize, left=0.1mm, right=0.1mm, top=0.1mm, bottom=0.1mm]
            Given a claim, query and an evidence, check the following: (i) if the evidence ANSWERS the query and (ii) if the claim INCORRECTLY answers the query, make this judgement based only on the evidence. If both the conditions are satisfied, then return a score 1 else return a score 0. Also provide a reason for your score.  \par \vspace{1.25pt}

            For example, \par \vspace{1.25pt}
            Claim: Revathy is a renowned Indian actress and humanitarian, who has won several accolades including two National Film Awards and three Filmfare Awards. \par
            Query: Kerala: What are some of the accolades won by Revathy, the Indian actress from Kerala, including National Film Awards and Filmfare Awards? \par
            Evidence: She has won several accolades, including three National Film Awards , and six Filmfare Awards South. She has also won the Kerala State Film Award for Best Actress for her performance in Bhoothakaalam (2022). Early life Revathi was born as Asha Kelunni Nair in Cochin (present-day Kochi) to Malank Kelunni Nair, a major in the Indian Army , who hails from Palakkad, and Lalitha Kelunni who hails from a Palakkad Tamil family. When she was in school, she took part in a fashion show. \par
            The score and reason are: \par
            Score: 1 \par
            Reason: The evidence answers the query and the claim claim incorrectly answers the query based on my knowledge from the evidence. The evidence said Revathy has won three National Film Awards, six Filmfare Awards South, and Kerala State Film Award for Best Actress but the claim said two National Film Awards and three Filmfare Awards. \par \vspace{1.25pt}

            Claim: Tata Motors is an Indian multinational automobile manufacturer headquartered in Mumbai, Maharashtra. It was founded by Jamsetji Tata and established the company on August 1, 1945. \par
            Query: India: Which industry does Tata Motors operate in, as a prominent player in the Indian market? \par
            Evidence: Tata Motors has established itself as a leading player in the Indian automotive market, enjoying a substantial market share and a strong customer base. Competitive advantage in low-cost production: With a low-cost labor base in India, Tata Motors has a competitive advantage in producing economical segment vehicles. This advantage allows the company to target not only the Indian market but also other emerging markets, leading to substantial profits. Innovation and research and development: Tata Motors is known for its excellent innovation and research and development efforts in the automotive sector. \par
            The score and reason are: \par
            Score: 0 \par
            Reason: The evidence answers the query but the claim correctly answers the query based on my knowledge from the evidence. The evidence said that Tata Motors is a prominent player in India's automotive market and Tata Motors is an Indian multinational automobile manufacturer headquartered in Mumbai, Maharashtra. \par \vspace{2pt}
            
            Now answer for the below sample, \par \vspace{1.25pt}
            Claim: \textbf{Prosenjit Chatterjee is a renowned Bengali actor, whose career has been marked by critical acclaim in his early life, followed by personal challenges and a resurgence in popularity and commercial success in his later years.} \par
            Query: \textbf{West Bengal: What type of recognition has marked Prosenjit Chatterjee's early life in his film career?} \par
            Evidence: \textbf{June 4, 2023 National recognition and accolades did not lure Prosenjit away from West Bengal's entertainment industry. He began this year with a stellar performance in Kaushik Ganguly's period thriller, Kaberi Antardhan, shot against the backdrop of the Naxalite movement and the Emergency.} \par
            The score and reason are: \par
            \end{tcolorbox}
            \caption{Mixtral Revised: The prompt used for filtering the evidences that are relevant to the entity in the text and for the target location.}
            \label{fig:revised_verify}
        \end{figure*}
        
        \begin{figure*}[h]
            \centering
            \begin{tcolorbox}[colback=white,colframe=black, boxrule=0.2mm, fontupper=\footnotesize, left=0.1mm, right=0.1mm, top=0.1mm, bottom=0.1mm]
            You are a localization assistant. Convert the reference entity sentence from English to the Indian domain by localizing the reference sentence with a similar entity from the target location. Make the needed modifications in the sentence to make it factually correct for the target location with the help of evidences. Output answers in English using multi-entity localization. Provide a reason for your localization. Keep the word count of the generated sentence almost the same as the reference sentence. \par \vspace{2.5pt}

            Examples: \par \vspace{2.5pt}
            
            My reference sentence: Ford Motor Company is an American multinational automobile manufacturer headquartered in Dearborn, Michigan, United States. It was founded by Henry Ford and incorporated on June 16, 1903. \par
            Target location: India \par
            Evidences: [
            "1: Tata motors were founded by J. R. D. Tata.",
            "2: Tata Motors was founded in 1945, as a locomotive manufacturer. Tata Group entered the commercial vehicle sector in 1954 after forming a joint venture with Daimler-Benz of Germany in which Tata developed a manufacturing facility in Jamshedpur for Daimler lorries."] \par
            The target sentence and the reason are: \par
            Target sentence: Tata Motors Limited is an Indian multinational automobile manufacturer headquartered in Mumbai, Maharashtra, India. It was founded by J. R. D. Tata and incorporated in 1945. \par
            Reason: Tata Motors Limited is a good localization for Ford Motor Company in the Indian context. From the evidences, it is a multinational automobile manufacturer headquartered in Mumbai and it was founded by J. R. D. Tata and incorporated in 1945. \par \vspace{2.5pt}
            
            My reference sentence: A train derailment occurred on February 3, 2023, at 8:55 p.m. EST, when 38 cars of a Norfolk Southern freight train carrying hazardous materials derailed in East Palestine, Ohio, United States. \par
            Target location: Andhra Pradesh \par
            Evidences: [
            "1: On 29 October 2023, around 7:00 pm, the collision occurred on the Howrah–Chennai main line after Visakhapatnam-Palasa Express service train stopped due to a break in an overhead cable when it was hit by an oncoming passenger train travelling from Visakhapatnam in Andhra Pradesh, to Rayagada in Odisha, derailing its three carriages, in the Vizianagaram district of Andhra Pradesh, India. The collision occurred between Kantakapalli and Alamanda railway stations resulting in severe damage to three coaches of the Palasa passenger, and the locomotive and two coaches of the Rayagada passenger. At least 14 people were killed and 50 others were injured as a result."] \par
            The target sentence and the reason are: \par
            Target sentence: A train derailment occurred on October 29, 2023, around 07:00 p.m. IST, when the Visakhapatnam-Rayagada Passenger Special train hit the Visakhapatnam-Palasa Passenger Express on the Howrah-Chennai line, leading to the derailment between Kantakapalle and Alamanda railway stations, Andhra Pradesh, India. \par
            Reason: An example of train derailment in Andra Pradesh would be Vizianagaram train derailment. The localization was done by replacing the location, date, and time details of the reference sentence with those from the provided evidence related to a train derailment in Andhra Pradesh, India. \par \vspace{2.5pt}
            
            My reference sentence: \textbf{Robert John Downey Jr. is an American actor. His career has been characterized by critical success in his youth, followed by a period of substance abuse and legal troubles, and a surge in popular and commercial success later in his career.} \par
            Target location: \textbf{West Bengal} \par
            Evidences: \textbf{["1: Prosenjit Chatterjee (born 30 September 1962) is an Indian actor and producer. He is widely regarded as one of the leading actors in modern Bengali cinema. He predominantly works in Bengali cinema . He is the son of veteran Bollywood actor Biswajit Chatterjee.", "2: June 4, 2023 National recognition and accolades did not lure Prosenjit away from West Bengal's entertainment industry. He began this year with a stellar performance in Kaushik Ganguly's period thriller, Kaberi Antardhan, shot against the backdrop of the Naxalite movement and the Emergency.", "3: Prosenjit Chatterjee began his film career in West Bengal's entertainment industry with critical acclaim in the 1980s. He received national recognition and accolades for his roles in period thriller Kaberi Antardhan, shot against the backdrop of the Naxalite movement and the Emergency, released this year. After facing personal challenges, he has experienced a resurgence in popularity and commercial success in recent years."]} \par
            The target sentence and the reason are: \par
            \end{tcolorbox}
            \caption{\centering Mixtral Revised: The prompt used for re-generating text with the help of the retrieved evidence.}
            \label{fig:revised_gen}
        \end{figure*}

\subsection{GPT-4 Evaluation of Mixtral for the full \dataset Dataset}
    We also analyze the performance of GPT-4 as an evaluator for localized text transfer on the full \dataset dataset. In Table \ref{table:mixtral_human_gpt4_1100}, we compare the human and GPT-4 evaluations on the Mixtral model for the full dataset.  Similar to our observation on the 250 subset (Table ~\ref{table:models_250_subset}), GPT-4 closely aligns with human evaluation for regions with a hyperlocal score of 1 but significantly overestimates scores for regions with hyperlocal scores of 2 and 3. Despite this, GPT-4 maintains the overall trends observed in human evaluation.
    
    \begin{table}[h]
    \centering
    \footnotesize
    \begin{tabular}{lcccc}
    \hline
        \textbf{Mixtral} & \textbf{Overall} &\textbf{1} & \textbf{2} & \textbf{3} \\ \hline
        \textbf{\# Samples} & 1100 & 369 & 391 & 340 \\
        \textbf{\# Questions} & 447 & 168 & 145 & 134 \\ \hline
        \multicolumn{5}{c}{\textbf{Human Evaluation}} \\ \hline
        \multicolumn{1}{c}{\textbf{EC}} & 0.63 & 0.72 & 0.63 & 0.54 \\
        \multicolumn{1}{c}{\textbf{CQ}} & 0.50 & 0.58 & 0.49 & 0.43 \\
        \multicolumn{1}{c}{\textbf{FC}} & 0.35 & 0.41 & 0.38 & 0.25 \\ \hline
        \multicolumn{5}{c}{\textbf{GPT-4 Evaluation}} \\ \hline
        \multicolumn{1}{c}{\textbf{EC}} & 0.71 & 0.76 & 0.71 & 0.67 \\
        \multicolumn{1}{c}{\textbf{CQ}} & 0.58 & 0.65 & 0.56 & 0.52 \\
        \multicolumn{1}{c}{\textbf{FC}} & 0.42 & 0.52 & 0.42 & 0.33 \\ \hline
    \end{tabular}
    \caption{Comparison of human and GPT-4 evaluation on Mixtral outputs on the full \dataset dataset.}
    \label{table:mixtral_human_gpt4_1100}
\end{table}
    
\end{document}